%% file: main.tex
\documentclass{article}
\usepackage[utf8]{inputenc}
\usepackage[margin=1.25in]{geometry}

\usepackage[colorlinks=true,citecolor=blue,urlcolor=blue]{hyperref}
\usepackage{booktabs}
\usepackage{amsmath,amsfonts,amsthm}
\usepackage{mathtools}
\usepackage{xcolor}
\usepackage{xspace}
\usepackage{nicefrac}

\usepackage{caption}
\usepackage{varwidth}

\usepackage[capitalise]{cleveref}
\Crefname{equation}{Eq.}{Eqs.}
\Crefname{figure}{Fig.}{Figs.}
\Crefname{table}{Tab.}{Tabs.}
\Crefname{algorithm}{Alg.}{Algs.}
\Crefname{section}{Sec.}{Secs.}
\Crefname{appendix}{App.}{Apps.}
\Crefname{theorem}{Thm.}{Thms.}
\Crefname{remark}{Rmk.}{Rmks.}

\usepackage{graphicx}
\graphicspath{./images/}
\usepackage[skip=0pt]{subcaption}

\usepackage[framemethod=tikz]{mdframed}
\usepackage{multicol}
\usepackage{multirow}
\usepackage[noend]{algorithmic}
\usepackage{algorithm}

\newcommand{\noisestate}{\mathbf{\mathcal{S}}}
\newcommand{\optstate}{\mathcal{P}}

\usepackage[inline]{enumitem}
\setenumerate[1]{itemsep=0pt,partopsep=0pt,parsep=\parskip,topsep=5pt}
\setitemize[1]{itemsep=0pt,partopsep=0pt,parsep=\parskip,topsep=5pt}
\setdescription{itemsep=0pt,partopsep=0pt,parsep=\parskip,topsep=5pt}

\input{macros.tex}
\newif\ifacl

\definecolor{darkgreen}{rgb}{0,0.4,0.0}

\PassOptionsToPackage{sort}{natbib}
\PassOptionsToPackage{square}{natbib}
\usepackage[numbers]{natbib}

\title{A Hassle-free Algorithm for Private Learning in Practice: \\ Don't Use Tree Aggregation, Use BLTs
}

\author{
H. Brendan McMahan,\;\;Zheng Xu,\;\;Yanxiang Zhang\thanks{Alphabetical order. \texttt{\{mcmahan, xuzheng, zhangyx\}@google.com}}\vspace{0.5em}\\
\textit{Google}\vspace{0.5em}\\
}
\date{}

\begin{document}
\sloppy
\setboolean{acl}{false} 

\maketitle

\input{sec_abs}

\input{sec_intro}

\input{sec_dpftrl}

\input{sec_blt_mp}

\input{sec_exp}

\input{sec_gboard}

\input{sec_concl}

\subsection*{Acknowledgement}
The authors thank Zachary Garrett, Nicole Mitchell, and Keith Rush for supporting the production code; Yu Xiao for helping some of the production experiments; Ryan McKenna and Yuanbo Zhang for reviewing an early draft; and Nikita Kalinin and Krishna Pillutla for pointing out typos in \cref{alg:multC} and \cref{alg:blt_pair} in the initial arXiv version.

\bibliographystyle{plainnat}
\bibliography{refer}

\clearpage

\appendix
\sloppy
\input{sec_app}

\end{document}

%% file: macros.tex
\usepackage{bm}

\DeclarePairedDelimiter\ceil{\lceil}{\rceil}

\newtheorem{lem}{Lemma}[section]
\newtheorem{theorem}[lem]{Theorem}

\newtheorem{rem}[lem]{Remark}
\newtheorem{prop}[lem]{Proposition}
\newtheorem{definition}[lem]{Definition}

\newcommand{\alg}[1]{\textsc{#1}\xspace}
\newcommand{\bandmf}{\alg{BandMF}}
\newcommand{\bandtoep}{\alg{BandToep}}
\newcommand{\blt}{\alg{BLT}}
\newcommand{\blts}{\alg{BLTs}}
\newcommand{\ourblt}{\alg{BLT$^\star$}}

\newcommand{\treeagg}{\alg{TreeAgg}}
\newcommand{\tafull}{\alg{TreeAgg-Full}}
\newcommand{\bandfhu}{\alg{BandFHU}}

\newcommand{\rmse}{RmsError\xspace}
\newcommand{\maxerr}{MaxError\xspace}
\newcommand{\rmsloss}{RmsLoss\xspace}
\newcommand{\maxloss}{MaxLoss\xspace}

\DeclareMathOperator{\rmseop}{RmsError}
\DeclareMathOperator{\maxerrop}{MaxError}
\DeclareMathOperator{\rmslossop}{RmsLoss}
\DeclareMathOperator{\maxlossop}{MaxLoss}
\DeclareMathOperator{\diag}{diag}

\newcommand{\LtToep}{\text{LtToep}}
\newcommand{\bltm}{\text{BLT}}

\newcommand{\hth}{\hat{\theta}}
\newcommand{\homega}{\hat{\omega}}

\newcommand{\STATEComment}{\STATE $\triangleright$\ }
\newcommand{\Comment}{\ $\triangleright$\ }
\newcommand{\Buf}{\bfS}  
\newcommand{\myind}{\hspace{0.5cm}}
\newcommand{\assign}{\leftarrow}

\DeclareMathOperator{\sens}{sens}
\newcommand{\abs}[1]{|#1|}

\newcommand{\bands}{\hat{b}}
\newcommand{\outp}{\omega} 

\newcommand{\obs}{\bfx}
\newcommand{\obsp}{\tilde{\bfx}}

\newcommand{\deltaset}{\mathfrak{D}}
\newcommand{\contrib}{\bfu}

\newcommand{\clipnorm}{\zeta}
\newcommand{\bparticipation}{$\minsep$-min-sep-participation\xspace}

\newcommand{\ind}{u}

\newcommand{\nd}{\ensuremath{n}} 

\newcommand{\nbuf}{d}  
\newcommand{\mdim}{m}  
\newcommand{\dimdim}{{\nd \times \nd}}

\newcommand{\maxpart}{k}  
\newcommand{\minsep}{b}
\newcommand{\inv}{^{-1}}

\newcommand{\bfA}{{\bm{A}}}
\newcommand{\bfB}{{\bm{B}}}
\newcommand{\bfC}{{\bm{C}}}

\newcommand{\bfI}{{\bm{I}}}

\newcommand{\bfN}{{\bm{N}}}

\newcommand{\bfS}{{\bm{S}}}

\newcommand{\bfX}{{\bm{X}}}

\newcommand{\bfZ}{{\bm{Z}}}

\newcommand{\bfu}{{\bm{u}}}

\newcommand{\bfx}{{\bm{x}}}

\newcommand{\norm}[1]{\left\|{#1}\right\|}
\newcommand{\tp}{^\top} 

\newcommand{\idx}[2]{_{#1, #2}}

\newcommand{\vfm}{:} 

\newcommand{\N}{\mathbb{N}}
\newcommand{\R}{\mathbb{R}}

\newcommand{\calO}{\mathcal{O}}

\newcommand{\var}[1]{\mathtt{#1}}
\newcommand{\hc}{\hat{c}}

%% file: sec_abs.tex
\begin{abstract}

The state-of-the-art for training on-device language models for mobile keyboard applications combines federated learning (FL) with differential privacy (DP) via the DP-Follow-the-Regularized-Leader (DP-FTRL) algorithm. Two variants of DP-FTRL are used in practice, tree aggregation and matrix factorization. However, tree aggregation suffers from significantly suboptimal privacy/utility tradeoffs, while matrix mechanisms require expensive optimization parameterized by hard-to-estimate-in-advance constants, and high runtime memory costs.
This paper extends the recently introduced Buffered Linear Toeplitz (BLT) mechanism to multi-participation scenarios. Our BLT-DP-FTRL maintains the ease-of-use advantages of tree aggregation, while essentially matching matrix factorization in terms of utility and privacy. 
We evaluate BLT-DP-FTRL on the StackOverflow dataset, serving as a re-producible simulation benchmark, and across four on-device language model tasks in a production FL system. Our empirical results highlight the advantages of the BLT mechanism and elevate the practicality and effectiveness of DP in real-world scenarios. 

\end{abstract}

%% file: sec_intro.tex
\section{Introduction}
Language models (LMs) that can predict the next word for input text are a powerful tool for many applications. In mobile keyboard applications, LMs are deployed on device to support various features (e.g., auto correction, smart completion and suggestion, and next word prediction) to improve users’ typing experience. On-device LMs are typically small (less than ten million parameters) due to latency requirement and limited on-device resources. Their performance can be significantly improved by training from user data~\citep{hard2018gboard,xu2023federated}; recent work~\citep{wang2023can,wu2024prompt} shows the necessity of training on user data to achieve high utility even when we can access large-scale web data and pre-trained large LMs with billions of parameters. 

As mobile-keyboard user data can be highly privacy sensitive, differential privacy (DP)~\citep{dwork2006calibrating,dwork2014algorithmic} and federated learning (FL)~\citep{mcmahan2017fedavg,kairouz2019advances} have emerged as best practices for such models. 
DP provides a mathematical formulation to upper-bound the memorization of an individual’s information in model training. FL minimizes data exposure by aggregating focused model updates from decentralized data stored only on user devices. DP and FL are combined when training on-device language models in production mobile keyboard applications\ifthenelse{\boolean{acl}}{~\citep{xu2023federated}}{~\citep{xu2023federated,choquette2023amplified,wu2024prompt}}. Applying DP in a production cross-device FL system is challenging as many DP algorithms require specific pattern of sampling training data to achieve strong privacy-utility trade-off. However, a cross-device FL system has limited control of sampling as clients can only participate in training when local criteria (e.g., charging, idle, and connected to an unmetered network) are satisfied~\citep{bonawitz2019towards,huba2022papaya}. 
Recently, DP-Follow-the-Regularized-Leader (DP-FTRL) algorithms~\citep{kairouz21practical,choquette2023amplified} have achieved superior privacy-utility trade-off with simpler client participation requirements, and are used in practice in FL systems~\citep{xu2023federated,zhang2023private}. 

Instead of requiring uniform or Poisson sampling of devices as in previous work~\citep{abadi2016deep,mcmahan2017learning}, DP-FTRL uses minimum separation (min-sep) to characterizes the participation pattern\ifthenelse{\boolean{acl}}{}{~\citep{kairouz21practical,xu2023federated}}. Min-sep is the smallest number of rounds between the consecutive participation of a client, and smaller min-sep necessitates adding more noise to achieve a desired DP guarantee. Min-sep is enforced in the FL system by implementing a timer on each device so that a device only becomes eligible for training if a certain period of time (e.g., three days) has passed since their last participation.
DP-FTRL algorithms leverage correlated noise mechanisms such as tree aggregation (\treeagg)~\citep{kairouz21practical} or matrix factorization (MF)~\citep{choquette2023amplified} with the client participation pattern in FL. The banded MF (\bandmf) mechanism\ifthenelse{\boolean{acl}}{}{~\citep{choquette2023amplified}} pre-computes matrices to generate  correlated noise from independent noise to achieve stronger DP guarantees than the \treeagg mechanism\ifthenelse{\boolean{acl}}{}{~\citep{kairouz21practical}}. \bandmf is superior when the number of rounds and min-sep can be (accurately) estimated before training to optimize matrices. However, min-sep is only known after training with time-based separation as many system factors may potentially affect training time~\footnote{Despite being more complicated in FL systems, it is possible to enforce round-based separation so that a device only becomes eligible for training if min-sep rounds has passed since their last participation. However, it is still challenging to pre-specify min-sep before training due to the dynamics of client availability and population size. If the target min-sep is too large, training might halt because of lacking eligible devices. If the target min-sep is too small, the MF mechanism is not optimal for the correlated noise generation.}.  Furthermore, \bandmf consumes more memory for noise generation, and hence is used less often than \treeagg in practice. 

In this work, we focus on the challenges of achieving strong DP guarantees in training production LMs in a cross-device FL system. We discuss how we extend the recent theoretical advancements of the Buffered Linear Toeplitz (\blt) mechanism from single participation~\citep{mcmahan2024efficient} to multi-participation scenarios, and adapt \blt to DP-FTRL. We apply \blt-DP-FTRL to FL in practice, demonstrating its advantages in flexibility, ease of use, and privacy-utility trade-offs. The \blt-DP-FTRL algorithm offers flexibility in handling varying numbers of training rounds, and robustness to a wide range of min separation between user participations. Furthermore, \blt-DP-FTRL simplifies the correlated noise generation process and reduces memory by exploiting the parameterization of the Toeplitz matrices. 
We empirically evaluate \blt-DP-FTRL on the StackOverflow benchmark dataset and across four on-device LM training tasks in a production FL system. Our \blt-DP-FTRL achieves better privacy-utility trade-off compared to the widely used \treeagg mechanism\ifthenelse{\boolean{acl}}{}{~\citep{kairouz21practical}}, and comparable results compared to the state-of-the-art \bandmf mechanism\ifthenelse{\boolean{acl}}{}{~\citep{choquette2023amplified}}. 
\blt-DP-FTRL exhibits desirable robustness properties in practice, offering a practical and effective solution for achieving strong DP in real-world FL systems.

%% file: sec_dpftrl.tex
\ifthenelse{\boolean{acl}}{\section{(\blt-)DP-FTRL for Private Learning}}{\section{(\blt-)DP-FTRL for Private Learning in Practice}} \label{sec:background}

\ifthenelse{\boolean{acl}}{\subsection{Background}}{}

\ifthenelse{\boolean{acl}}{
We use ($\epsilon$, $\delta$)-DP~\citep{dwork2006calibrating,dwork2014algorithmic} and $\rho$-zCDP (zero-Concentrated DP)~\citep{bun2016concentrated} to quantify the privacy protection: smaller $\epsilon$ ($\rho$) correspond to stronger DP guarantees. A formal definition and more discussion are in \cref{subsec:dp_def}.
}{
\input{subsec_dp_def}
\input{alg_dpfl}
}

\ifthenelse{\boolean{acl}}{\textbf{FL with DP}}{\subsection{Federated Learning with Differential Privacy (FL with DP)}}
We apply the generalized Federated Averaging (FedAvg) algorithm~\citep{mcmahan2017fedavg,wang2021fieldguide}, as shown in \cref{algo:dpfl} \ifthenelse{\boolean{acl}}{of \cref{app:background}}{}. 
FedAvg is the most common algorithm in cross-device FL systems. In a training round $t$ of total $n$ rounds, the server broadcasts a global model $y^t$ to a subset of clients; each client $i$ then updates their local model $y_i$ by SGD, and sends back the model delta; the model deltas are aggregated and used as a pseudo gradient on the server to update the global model. DP is achieved by clipping the $l_2$ norm of the model delta to control the sensitivity (contribution of each device), and then adding noise to the aggregated deltas on the server.

While our primary focus is federated learning with decentralized data in this paper, \cref{algo:dpfl} can also be applied in datacenter to achieve user-level DP~\citep{xu2022learning,chua2024mind,charles2024fine}. When using only one batch of a single sample for gradient computation in the ClientUpdate function and \treeagg for correlated noise, \cref{algo:dpfl} coincides with the DP-FTRL algorithm described in~\citep{kairouz21practical}. The DP guarantee is determined by noise calibrated to sensitivity, which depends on clip norm noise multiplier $\sigma$, the correlated noise mechanism, total number of rounds $T$, and client participation pattern (min-sep $b$). Clip norm $\zeta$ and clip norm noise multiplier $\sigma$ are used as algorithmic hyperparameters, similar to independent noise mechanism (e.g., DP-SGD/DP-FedAvg~\citep{abadi2016deep,mcmahan18learning}). However, instead of directly applying independent Gaussian noise of standard deviation $\sigma \zeta$, correlated noise are generated to privatize model updates. 

\ifthenelse{\boolean{acl}}{\textbf{MF for DP-FTRL}}{\subsection{Matrix Factorization (MF) for DP-FTRL}}
DP-FTRL~\citep{kairouz21practical} adds correlated noise to achieve strong privacy-utility trade-offs, observing that privatizing the prefix sum of model updates are essential for privatizing the training process.
The intuition of privatizing prefix sum is easier to understand when server optimizer is SGD, as the iterative process of per-round updates is equivalent to updating with prefix sum, i.e.,  
$$y^t=y^{t-1} - \eta_s \Delta^t = y^{-1}- \eta_s \sum_{j=0}^{t} \Delta^j.$$ We can write similar formulation for additional linear operation in optimization, such as momentum in SGD. In practice, it is often easier to get privatized per-round update by (conceptually) subtracting the privatized prefix sum from two consecutive rounds, and use the privatized update $\tilde{\Delta^t}$ in various server optimizers, guaranteed by the post-processing property of DP. 

We represent the model updates for $n$ rounds as a matrix $\bfX \in \R^{n \times \mdim}$, where each row is the sum of clipped updates (i.e., $\bfX_{t,\vfm} \coloneqq \sum_{i \in \mathcal{Q}^t} \Delta_i^t \in \R^{m}$ from~\cref{algo:dpfl}),
we aim to privatize $\bfA \bfX$, where $\bfA \in \R^{n \times n}$ is a lower triangular matrix of ones, i.e., $\bfA_{i,j}=1, \forall i\leq j$ and $\bfA_{i,j}=0, \forall i > j$. Given the privatized prefix sum $\widetilde{AX}$, the privatized model update is $\tilde{\Delta}^t \gets \widetilde{\bfA \bfX}_{t,\vfm} - \widetilde{\bfA \bfX}_{t-1,\vfm}$, and \cref{algo:dpfl} is privatized because of the post-processing property of DP. \citet{kairouz21practical} adopts the \treeagg\ifthenelse{\boolean{acl}}{}{\citep{dwork10continual,chan11continual,honaker15efficient}} mechanism to privatize $\bfA \bfX$. Recent work suggest a general matrix factorization framework\ifthenelse{\boolean{acl}}{~\citep{choquette2023amplified}}{~\citep{denisov22matfact, choquette2023amplified}} can be used to achieve even stronger privacy-utility trade-offs, and both \treeagg-DP-FTRL and the popular DP-SGD algorithm~\citep{abadi2016deep} are special cases in the MF-DP-FTRL framework. MF mechanism considers the factorization of $\bfA=\bfB \bfC$ and privatizes $\bfC \bfX$ by adding independent noise $\bfZ \in \R^{n \times m}$ with standard deviation $\sigma\zeta$.  
We can use the (pseudo-)inverse of the $\bfC$ matrix to generate the correlated noise in the streaming setting~\citep{choquette2023amplified},
\ifthenelse{\boolean{acl}}{
\begin{align}
\widetilde{\bfA \bfX} & = \bfB(\bfC \bfX + \zeta \bfZ) = \bfA \bfX + \zeta \bfC^{-1} \bfZ \nonumber \\ \Rightarrow \tilde{\Delta}^{t} & = \Delta^{t} + (\zeta \bfC^{-1} \bfZ)_{t, \vfm} \label{eq:matrixmech}
\end{align}
}{
\begin{equation}\label{eq:matrixmech}
\widetilde{\bfA \bfX} = \bfB(\bfC \bfX + \zeta \bfZ) = \bfA (\bfX + \zeta \bfC^{-1} \bfZ) \Rightarrow \tilde{\Delta}^{t} = \Delta^{t} + (\zeta \bfC^{-1} \bfZ)_{t, \vfm}
\end{equation}
}
\cref{eq:matrixmech} also suggests the alternative interpretation of correlated noise in DP-FTRL: at round t, the noise added in previous rounds can be cancelled when $\bfC^{-1}\idx{t}{\vfm}$ is negative.   

\ifthenelse{\boolean{acl}}{
\treeagg can be written in MF form, and the stronger variance reduction variant (\tafull)~\citep{honaker15efficient} is equivalent to setting $\bfB=\bfA \bfC^{-1} \in \R^{2^{l-1} \times (2^l-1)}$ by computing the Moore-Penrose pseudoinverse of $\bfC$~\citep{denisov22matfact}. However, the MF-based\tafull does not have a memory-efficient implementation, and consumes $n m$ memory. In this paper, we consider memory-efficient \treeagg~\citep{kairouz21practical} that is widely used in industry~\citep{xu2023federated}, and \tafull~\citep{denisov22matfact} that achieves better privacy-utility trade-off but less memory and computation efficient. \bandmf~\citep{choquette2023amplified} is the state-of-the-art for FL, which optimizes matrices with estimated min-sep bands. More related work  with detailed discussion are in \cref{app:ta_mf}.
}{
\input{subsec_ta_mf}
}

\subsection{\blt Mechanisms in DP-FTRL} \label{sec:background_blt}
We now consider lower-triangular Toeplitz matrix in MF mechanism, i.e., $\bfC := \LtToep(c) \in \R^\dimdim$ where $\bfC\idx{i}{j}=c_{i-j}, \forall i\leq j$ otherwise $C\idx{i}{j}=0$. Buffered-linear Toeplitz (\blt) matrices \citep{mcmahan2024efficient} parameterize $\bfC$ by $\theta \in (0, 1]^\nbuf$ (the ``buffer decay'' parameters) and non-negative $\omega \in \R_+^\nbuf$ (the ``output scale'' parameters), where the Toeplitz coefficients are given by
\begin{equation} \label{eq:closedcoefs}
c_i = 
\begin{cases}
1  & i = 0 \\
\sum_{j \in [\nbuf]} \omega_j \theta_j^{i-1} & i > 0.
\end{cases}
\end{equation}

The $\bltm(\omega, \theta)$ matrices have many useful properties\ifthenelse{\boolean{acl}}{}{~\citep{mcmahan2024efficient}}, most importantly for our purposes:
\begin{enumerate*}[label=\color{purple}(\arabic*)]
    \item Streaming multiplication by $\bfC$ ($\bfZ = \bfC \hat\bfZ$ for $\bfZ,\hat\bfZ \in \R^{\nd \times \mdim}$) can be computed efficiently using only $\calO(\nbuf \mdim)$ memory and $\calO(\nbuf \mdim)$ time per round $t$, without fully materializing $\bfC$, $\bfZ$, or $\hat\bfZ$. Hence $\bfC$ is referred as a $\nbuf$-buffer \blt.
    \item The inverse of a $\nbuf$-buffer \blt ($\bfC = \bltm(\omega, \theta)$) is another $\nbuf$-buffer \blt ($\bfC\inv = \bltm(\homega, \hth)$), and we can efficiently compute Toeplitz coefficients of $\bfC\inv$ using \cref{eq:closedcoefs} applied to $(\homega, \hth)$.
\end{enumerate*}
We now derive the correlated noise generation schema for $\bfC\inv\bfZ$ in \cref{eq:matrixmech} based on the BLT properties. We can first derive the BLT parameters $(\hth, \homega)$ of $\bfC\inv$ such that $\bfC\inv = \bltm(\hth, \homega)$, and then generate the correlated noise based on $(\hth, \homega)$ in streaming setting. However, we show a simpler alternative that directly uses the BLT parameters $(\theta, \omega)$ to generate streaming correlated noise $\hat\bfZ$. 

Applying the parameterization in \cref{eq:closedcoefs} \ifthenelse{\boolean{acl}}{}{(also appear in \citep[Sec 5]{mcmahan2024efficient})} to \citep[Alg 1]{mcmahan2024efficient}, and initializing buffers $\bfS_{-1} \gets \mathbf{0} \in \R^{\nbuf \times \mdim}$, we can efficiently compute $\bfZ_{t,\vfm}$ from $\hat\bfZ_{t,\vfm}$ in the streaming setting,
  $
  \bfZ_{t,\vfm} = \hat\bfZ_{t,\vfm} + \outp^T \bfS_{t-1}, 
  \bfS_{t} = \diag(\theta) \bfS_{t-1}  + \bm{1}_\nbuf \hat\bfZ_{t,\vfm}
  $. We rearrange the update equations to get $\hat\bfZ$ from $\bfZ$ and $\bfS$, 
  \ifthenelse{\boolean{acl}}{
  \begin{align}
  \hat\bfZ_{t,\vfm}  = \bfZ_{t,\vfm} - \outp^T \bfS_{t-1}, \nonumber \\ 
  \bfS_{t} = \diag(\theta) \bfS_{t-1}  + \bm{1}_\nbuf \hat\bfZ_{t,\vfm}. \label{eq:blt_noise}
  \end{align}
  }{
  \begin{align}
  \hat\bfZ_{t,\vfm}  = \bfZ_{t,\vfm} - \outp^T \bfS_{t-1},  
  \bfS_{t} = \diag(\theta) \bfS_{t-1}  + \bm{1}_\nbuf \hat\bfZ_{t,\vfm}. \label{eq:blt_noise}
  \end{align}
  }
  To efficiently generate correlated noise $\hat\bfZ_{t,\vfm}$ at round $t$, we only need to materialize the independent noise $\bfZ_{t,\vfm} \in \R^{1 \times m}$ and use the buffers $\bfS_{t-1} \in \R^{\nbuf \times \mdim}$ in \cref{eq:blt_noise}. The efficient correlated noise generation parameterized by BLT parameters $(\theta, \omega)$ for $\bfC\inv$ (instead of $\bfC$) did not appear in \citet{mcmahan2024efficient} and is new to this work. \cref{eq:blt_noise} intuitively shows the noise cancellation view of correlated noise, where the previous noises are tracked in the states decaying with $\theta \in (0, 1]$, and then subtracted in the current round after scaling with $\omega$. 
  
  For completeness, we provide the streaming multiplication algorithm of $\bfZ = \bfC \hat \bfZ$ and $\hat \bfZ = C^{-1} \bfZ$ for $\bfC = \bltm(\theta, \omega)$ in \cref{alg:multC} and \cref{alg:multCinv}, respectively. 
  \cref{alg:multC} is a direct application of \citep[Alg 1]{mcmahan2024efficient} and only used to derive \cref{alg:multCinv}, which is our streaming algorithm for generating correlated noise with \blts using only $dm$ memory. 
  Finally, we apply the streaming correlated noise $\hat\bfZ$ by BLT from \cref{alg:multCinv} (using \cref{eq:blt_noise})  in \cref{algo:dpfl} for the BLT-DP-FTRL algorithm, i.e., 
  \begin{equation}
      \tilde{\Delta}^t \leftarrow \sum_{i \in \mathcal{Q}^t} \Delta^{t}_i + \hat\bfZ_{t,\vfm}.
  \end{equation}

\ifthenelse{\boolean{acl}}{}{\input{alg_stream_multi}}

\ifthenelse{\boolean{acl}}{}{\input{tab_mech}}

%% file: subsec_dp_def.tex
\subsection{DP Formulation} \label{subsec:dp_def}
We present the definition of ($\epsilon$, $\delta$)-DP~\citep{dwork2006calibrating,dwork2014algorithmic} to quantify the privacy protection. 
\begin{definition}[$(\epsilon,\delta)$-Differential Privacy]
  A randomized algorithm $\mathcal{M}$ satisfies  ($\epsilon$, $\delta$)-DP for $\mathbb{D}$ if for any two neighboring datasets $\mathbb{D}$, $\mathbb{D}'$ and for all $\mathcal{S}\subset \text{Range}(\mathcal{M})$: 
\[
\textstyle{\Pr[\mathcal{M}(\mathbb{D}) \in \mathcal{S}] \leq e^{\epsilon}\Pr[\mathcal{M}(\mathbb{D}') \in \mathcal{S}] +\delta}
.\]
\end{definition}
Smaller ($\epsilon$, $\delta$) values suggest stronger DP guarantees, and we often measure $\epsilon$ at a fixed small $\delta=10^{-10}$. DP-FTRL also uses an alternative definition, $\rho$-zCDP (zero-Concentrated DP)~\citep{bun2016concentrated} designed for Gaussian mechanism, and smaller $\rho$ suggests stronger DP guarantees. We use PLD (privacy Loss Distributions) accounting~\citep{doroshenko2022connect,pldlib} to convert $\rho$-zCDP to ($\epsilon$, $\delta$) DP. When applying DP in FL, the neighboring datasets $\mathbb{D}$, $\mathbb{D}'$ are defined by zeroing out the contribution of all data on a user device. More discussions on neighboring dataset for the streaming setting in learning, and connection to DP guarantees are provided in \cref{sec:sensitivity}.

%% file: alg_dpfl.tex
\begin{algorithm}[thb]
\caption{\small FedAvg~\citep{mcmahan18learning} with \colorbox{blue!25}{DP-FTRL~\citep{kairouz21practical}} for \colorbox{blue!25}{DP} FL}
\label{algo:dpfl}
\begin{algorithmic}
\INPUT: clients per round $m$, learning rate on client $\eta_{c}$ and on server $\eta_{s}$, momentum $\beta=0.9$,  total number of rounds $T$, \colorbox{blue!25}{clip norm $\zeta$, clip norm noise multiplier $\sigma$,} 
\begin{multicols}{2}
\STATE Initialize model $y^{-1}$ with pretraining
\STATE Initialize server optimizer state $\optstate$
\STATE \colorbox{blue!25}{Initialize correlated noise state $\noisestate$ with $\sigma \zeta$}
\FOR{each round $t=0, 1, 2, \ldots, n-1$}
\STATE $\mathcal{Q}^t \leftarrow$ (at least $m$ users that did not participate in the previous $b$ rounds)
\FOR{each user $i \in \mathcal{Q}^t$ \textbf{in parallel}}
\STATE $\Delta^{t}_i \leftarrow$ ClientUpdate$(i, y^{t-1})$
\ENDFOR
\vspace{0.1cm}
\STATE \colorbox{blue!25}{$\tilde{\Delta}^t, \noisestate \leftarrow$AddCorrNoise$(\noisestate, \sum_{i \in \mathcal{Q}^t} \Delta^{t}_i)$}
\vspace{0.1cm}
\STATE $y^{t}, \optstate \leftarrow$ServerOpt$(y^{t-1}, \frac{1}{m}\tilde{\Delta}^t, \eta_{s}, \beta, \optstate)$ 
\ENDFOR

\begin{mdframed}[
    linecolor=black,
    linewidth=1.5pt,
    roundcorner=4pt,
    userdefinedwidth=\linewidth,
]
\FUNCTION{ClientUpdate($i$, $x_i$)}
\STATE $\mathcal{G} \leftarrow $ (batches of user $i$'s local data)
\FOR{batch $g \in \mathcal{G}$}
\STATE $y_i \leftarrow y_i - \eta_c \nabla \ell(y_i;g)$
\ENDFOR
\STATE $\Delta \leftarrow y_i - y_i^{(0)}$
\STATE \colorbox{blue!25}{$\Delta' \leftarrow \Delta \cdot \min{\left(1, \frac{\zeta}{||\Delta||}\right)}$}
\STATE \textbf{return} $\Delta'$
\ENDFUNCTION
\end{mdframed}
\end{multicols}
\end{algorithmic}
\vspace{-0.2cm}
\end{algorithm}

%% file: subsec_ta_mf.tex
\paragraph{\treeagg} can be written in MF form by recursively defining $\bfC^l \in \{0 ,1\}^{(2^l-1) \times 2^{l-1}}$ $l=\ceil{\log_2 n}$, as $\bfC^1 = [1]$, $\bfC^l = [[\bfC^{l-1}, \bm{0}], [\bm{0}, \bfC^{l-1}], [\bm{1}, \bm{1}]]$, where each row $\bfC_{i,\vfm}$ represents a node in the binary tree and the ones in $\bfC_{i,\vfm}$ represent the leaf nodes for a subtree. After adding noise $\bfZ$ to every tree node, vanilla \treeagg uses matrix $\bfB$ to selects and aggregates tree nodes to privatize the prefix sum, i.e., $\bfB \in \{0, 1\}^{2^{l-1} \times (2^l-1) }$ has $\bfB_{i,j}=1, \forall j=2^{k+1}-1, k \in \kappa, i = \sum_{k \in \kappa} 2^k,$ otherwise $\bfB_{i,j}=0$. Several schemes improve vanilla binary \treeagg for prefix sums appear in the literature.  \citet{kairouz21practical} efficiently implemented \treeagg with partial variance reduction~\citep{honaker15efficient}, which leverages the recursive structure of $C$ and only needs $\ceil{\log_2 n}m$ memory to generate correlated noise. The full variance reduction trick~\citep{honaker15efficient} can further improve the performance and is equivalent to setting $\bfB=\bfA \bfC^{-1} \in \R^{2^{l-1} \times (2^l-1)}$ by computing the Moore-Penrose pseudoinverse of $\bfC$~\citep{denisov22matfact} (we use an abuse of notation $\bfC^{-1}$ for pseudoinverse). However, the full variance reduction \treeagg (\tafull) does not have a memory-efficient implementation, and consumes $n m$ memory. Another variant~\citep{andersson24smooth} is more memory-efficient, but achieves suboptimal performance compared to MF approaches~\citep{fichtenberger2022constant}. In this paper, we primarily consider \treeagg~\citep{kairouz21practical} that is widely used in industry~\citep{xu2023federated}, and \tafull~\citep{denisov22matfact} that achieves better privacy-utility trade-off but less memory and computation efficient, to represent the tree aggregation mechanisms.

\paragraph{\bandmf}
\citet{choquette2023amplified} exploits the banded structure, i.e., $\bfC \in \R^{n\times n}$ where $\bfC_{i,j}=0, \forall |i-j| \geq \hat{b}$, to simplify the optimization and privacy accounting for MF mechanisms. \bandmf successfully applied MF mechanisms to the FL system for the first time. When fixing all the other configurations for training a production LM, \bandmf improved the DP guarantee from $\rho=0.52$-zCDP by \treeagg to $\rho=0.24$-zCDP. However, \bandmf has to estimate the band size $\hat{b}$ and total rounds $n$ for optimizing matrices before training, and the performance quickly drops when the actual min-sep $b$ in FL training is smaller than $\hat{b}$, or the training round is more than $n$. \bandmf improves memory usage of MF from $n\times m$ to $\hat{b} \times m$ for correlated noise, but the typical value of min-sep $b$ in FL is still hundreds to thousands for strong DP guarantees. More recently, \bandfhu~\citep{kalinin2024banded} and \bandtoep~\citep{mckenna2024scaling} optimize banded Toeplitz matrices for larger $\nd$ and exploit Toeplitz structure for computation efficiency, but they have not been shown to outperform \bandmf in the FL setting.

%% file: alg_stream_multi.tex
 \begin{minipage}[t]{0.49\textwidth}
\begin{algorithm}[H]
\caption{\small Stream Mult. by $\bltm(\theta, \omega)$~\citep{mcmahan2024efficient}} \label{alg:multC}
\begin{algorithmic}
\STATE \textbf{Input:} 
\STATE \myind Input stream $\hat \bfZ \in \R^{\nd \times \mdim}$
\STATE \myind $\theta \in \R^\nbuf, \omega \in \R^\nbuf$ for $\bfC = \bltm(\theta, \omega)$
\STATE \textbf{Output:}
\STATE \myind The rows $\bfZ\idx{t}{:}$ of $\bfZ = \bfC \hat \bfZ$
\STATE
\STATE Initialize buffers $\Buf_{-1} \assign \mathbf{0} \in \R^{\nbuf \times \mdim}$
\FOR{$t=0, \dots, n - 1$}
  \STATE  $\bfZ_{t,\vfm} = \hat\bfZ_{t,\vfm} + \outp^T \bfS_{t-1}$ 
  \STATEComment Decay each buffer by $\theta$ and add $\hat\bfZ_{t,\vfm}$ to each
  \STATE $\bfS_{t} = \diag(\theta) \bfS_{t-1}  + \bm{1}_\nbuf \hat\bfZ_{t,\vfm}$
  \STATE Output $\bfZ_{t,\vfm}$
\ENDFOR
\end{algorithmic}
\end{algorithm}
\end{minipage}
\hfill
\begin{minipage}[t]{0.49\textwidth}
\begin{algorithm}[H]
\caption{\small Stream Mult. by $\bltm^{-1}(\theta, \omega)$} \label{alg:multCinv}
\begin{algorithmic}
\STATE \textbf{Input:} 
\STATE \myind Input stream $\bfZ \in \R^{\nd \times \mdim}$
\STATE \myind $\theta \in \R^\nbuf, \omega \in \R^\nbuf$ for $\bfC = \bltm(\theta, \omega)$
\STATE \textbf{Output:}
\STATE \myind The rows $\hat \bfZ\idx{t}{:}$ of $\hat \bfZ = \bfC^{-1} \bfZ$

\STATE
\STATE Initialize buffers $\Buf_{-1} \assign \mathbf{0} \in \R^{\nbuf \times \mdim}$
\FOR{$t=0, \dots, n - 1$}
  \STATE $\hat\bfZ_{t,\vfm}  = \bfZ_{t,\vfm} - \outp^T \bfS_{t-1}$
  \STATEComment The buffer update is the same as \cref{alg:multC}
  \STATE $  \bfS_{t} = \diag(\theta) \bfS_{t-1}  + \bm{1}_\nbuf \hat\bfZ_{t,\vfm}$
  \STATE Output $\hat \bfZ\idx{t}{:}$
\ENDFOR
\end{algorithmic}
\end{algorithm}
\end{minipage}

%% file: tab_mech.tex
\begin{table}[t]
\renewcommand{\arraystretch}{1.4}
\centering
\small
\begin{tabular}{p{0.9in} p{0.7in} p{0.8in} p{0.9in} p{0.5in} p{0.4in}}
\toprule
Mech & Loss & Mech. opt. \newline cost & Memory \newline overhead  & Noise Added & $(n,\minsep)$-fragility \\
\midrule
\blt (ours)
  & \maxloss/ \rmsloss
  & Low \newline ($\calO(n)$ )  
  & Low \newline ($\sim 4 \times \mdim$  )         
  & Low
  & Low \\
\bandmf\ifthenelse{\boolean{acl}}{}{\citep{choquette2023amplified}}
  & \rmsloss
  & High \newline ($\calO(n^2)$)
  & High \newline ($\bands \times \mdim$ )     
  & Lowest 
  & Med \\
\bandtoep\ifthenelse{\boolean{acl}}{}{\citep{mckenna2024scaling}}
  & \rmsloss
  & Low  \newline ($\calO(n)$)
  & High \newline ($\bands \times \mdim$)
  & Low 
  & Med \\
\treeagg\ifthenelse{\boolean{acl}}{}{\citep{kairouz21practical}}  
  & \rmsloss\!\!*
  & Lowest \newline (predefined)
  & Med \newline  ($\ceil{\log_2(\nd)} \times \mdim$) 
  & High 
  & Low  \\
\bottomrule
\end{tabular}
\caption{\small Summary of mechanisms considered, evaluated in terms of: \emph{mechanism optimization cost}, how expensive is it to compute the mechanism $\bfC$; $\calO(\cdot)$ gives the cost of a single gradient calculation. The next two columns relate to the deployment of the mechanism in a ML training system: \emph{memory overhead} is the additional state (as a multiple of the model dimension $\mdim$) that the server needs to maintain; the per-round runtime cost is also proportional to this value. The \emph{noise added} is categorized subjectively, for details see examples in ~\cref{fig:rmse}. $(n, \minsep)$-fragility reflects the degree to which the mechanisms performance degrades when total number of rounds $n$ and min-sep $\minsep$ are poorly estimated, see discussion on quantitative results in \cref{fig:rmse_k3,fig:rmse_nkb,fig:rmse_vary_n,fig:other_mechs}.  The conclusion:  \textbf{\blts perform well on all aspects.}
} \label{tab:mechanism_summary}
\end{table}

\subsection{Comparing DP-FTRL Mechanisms}
We summarize DP-FTRL mechanisms in \cref{tab:mechanism_summary} \ifthenelse{\boolean{acl}}{in \cref{app:blt-dp-ftrl}}{} and show the advantages of BLT-DP-FTRL. Our BLT mechanism can optimize either MaxLoss or RmsLoss for generating correlated noise (detailed in \cref{sec:opt_blt} following~\citep{mcmahan2024efficient}), while previous MF mechanisms in practice primarily consider RmsLoss~\citep{choquette2023amplified,mckenna2024scaling}. It is possible to extend the previous mechanisms to use MaxLoss, while it is still an open problem which loss format corresponds better with learning performance when running with the DP-FTRL algorithms. \treeagg, especially \tafull, is equivalent to considering RmsLoss though the mechanism is predefined without explicit optimization cost; and we present the lower memory overhead ($\ceil{\log_2(\nd)} \times \mdim$) for \treeagg while \tafull without an efficient algorithm yet actually needs . 

BLTs achieve better privacy-utility trade-offs than \tafull in simulation benchmark experiments (see \cref{sec:exp}), and clearly outperforms \treeagg in production cross-device FL experiments (see \cref{sec:gboard}), as lower noise is added in BLTs. While \bandmf\citep{choquette2023amplified} can add lowest noise (measured by RmsLoss in \cref{fig:rmse}), BLTs have lower mechanism optimization cost and memory overhead. Moreover, \cref{sec:exp,sec:gboard} show the learning performance of BLTs are often comparable with \bandmf under the same privacy guarantee in practical settings, though BLTs' RmsLoss is slightly worse. The memory overhead of BLTs is $d\times m$ where we empirically observe that buffer size $d=4$ achieves low losses and further increasing $d$ does not further improve in our empirical settings of total rounds $n$ and min-sep $b$. The BLT memory overhead of $d \sim 4$ is smaller than \treeagg where $\ceil{\log_2(n)} \sim 11$, and much smaller than typical $\hat b \sim X00$ for \bandmf and \bandtoep. \bandtoep~\citep{mckenna2024scaling} suggested small $\hat b$ is preferred when using amplification by sampling in the many participation settings; however, sampling is generally not possible in practical cross-device FL systems. 

As shown in \cref{fig:rmse_k3,fig:rmse_nkb,fig:rmse_vary_n,fig:other_mechs}, BLT is also more robust than banded MF mechanisms when number of total rounds $n$ and min-sep $b$ are not accurately estimated. Specifically, it is unclear how to run Banded MF mechanisms beyond the estimated $n$ after optimizing the $\bfC \in \R^{n, n}$ matrix for correlated noise. Optimizing $\bfC \in \R^{n, n}$ for a much larger $n$ and truncating it to the actual number of training rounds can achieve good privacy-utility trade-offs, but encounter non-trivial mechanism optimization cost. BandMF performance degrades fast when the actual min-sep $b$ is smaller than the estimated band $\hat b$, but the stronger DP guarantees are generally achieved when $\hat b$ is large. Hence the tricky task of estimating min-sep $b$ is more important for \bandmf. In general, BLT-DP-FTRL is competitive for achieving state-of-the-art privacy-utility trade-off (compared to \bandmf), while maintains ease to use in practice (compared to \treeagg).  

%% file: sec_blt_mp.tex
\ifthenelse{\boolean{acl}}{\section{Multi-participation \blts}}{\section{Optimizing \blts for Multiple Participations}} \label{sec:blt_mp}
We study how to optimize for the BLT parameters $\theta \in \R^d$ and $\omega \in \R^d$ in \cref{eq:blt_noise} for the BLT-DP-FTRL algorithm, and account for DP guarantees. Particularly, we generalize the BLT optimization and DP accounting in \citep{mcmahan2024efficient} from single participation to multiple participations

\subsection{Sensitivity Under Multiple Participations}\label{sec:sensitivity}

\ifthenelse{\boolean{acl}}{
We provide additional background about multi-participation sensitivity definition, computation and usage in DP in \cref{app:mp_sens}, and only discussing the main results in this section. We further derive a lower bound for sensitivity in \cref{app:sens_lower_bound} used for \treeagg in simulation experiments in \cref{sec:exp}.
}{
\input{subsec_blt_sens}

}

\newcommand{\cc}{\mathbf{c}}

\ifthenelse{\boolean{acl}}{}{\paragraph{Multi-participation Sensitivity for \blts}}
Let $\bfC = \LtToep(c) \in \R^\dimdim$ be a lower-triangular Toeplitz matrix defined by the sequence of Toeplitz coefficients $\cc = (c_0, c_1, \dots, c_{\nd-1}) \in \R^\N$ as in \cref{sec:background_blt}. We assume $c_i \ge 0$ and $\cc$ is non-increasing, and consider the sensitivity of $\bfC$. 
Let $\cc_i = \bfC\idx{:}{i}$ be the $i$th column of $\bfC$, so $\cc_0 = \cc$ and generally 
\[
\cc_j = \big(\underbrace{0, 0, \dots, 0}_{\text{$j$ zeros}}, c_0, c_1, \dots c_{n-j-1}\big) \in \R^\nd.
\]

\newcommand{\pistar}{\pi^\star}
\newcommand{\ustar}{\ind(\pistar)}
\newcommand{\ii}{{\tilde{\imath}}}
\newcommand{\kk}{k}

The sensitivity of general Toeplitz matrices with decaying coefficients is recently discussed in \citet[Thm. 2]{kalinin2024banded}, which we restate in \cref{thm:toepminsepsens} with our notation. The participation pattern $\pistar$ simply puts the $\maxpart$ participations as early as possible, with each participation separated by exactly $\minsep$. This sensitivity computation is important for both DP accounting and optimizing for BLT parameters in \cref{sec:opt_blt}. 

\begin{theorem}\label{thm:toepminsepsens}
Given a Toeplitz strategy matrix $\bfC = \LtToep(\cc) \in \R^\dimdim$ with $\cc$ non-increasing and non-negative. Then, $\sens_{\Pi_\minsep}(\bfC)$ can be computed in time $\calO(\maxpart \nd)$  as
\[
  \sens_{\bfN_\Pi}(\bfC) = \norm{\bfC \ustar}_2
\]
where $\pistar$ is given by
\begin{equation}\label{eq:badpi}
\pistar = (0, b, 2b, \dots, (k-1)b).
\end{equation}
\end{theorem}

\ifthenelse{\boolean{acl}}{
}{
\input{subsec_sens_lower_bound}

}

\ifthenelse{\boolean{acl}}{\subsection{Analytical Utility as Objective}}{\subsection{Evaluating Matrix Factorization Mechanisms}}\label{sec:mf_obj}

While our end goal is good learning performance (as measured by held-out test set accuracy), we can estimate the utility of a matrix mechanism for DP-FTRL by quantifying the error it introduces into prefix sum estimates. 
The total noise introduced by the DP mechanism into prefix sum estimates in \cref{eq:matrixmech} will be $\bfB \bfZ = \widetilde{\bfA \bfX} - \bfA \bfX$ where $\bfZ \in \R^{\nd \times \mdim}$ is IID Gaussian noise with $\sigma$ determined according to the desired DP guarantee, and $\bfB = \bfA \bfC\inv$. We consider two error metrics based on the standard deviation of the total noise added to the prefix sum estimates.  The \maxerr is the worst-case standard deviation in the estimate of any prefix sum, which can be computed as %
{\small %
\[
\maxerrop(\bfB) \coloneqq \max_{i \in [\nd]} \sqrt{\sum_{j \in [\nd]} \bfB_{i,j}^2};
\]
}%
similarly the root-mean-squared error over all iterations $i \in [\nd]$ is %
{\small %
\[
\rmseop(\bfB) \coloneqq 
\sqrt{\sum_{i \in [\nd]} \sum_{j \in [\nd]} \bfB_{i,j}^2 / \nd}.
\]
} %

The standard deviation $\sigma$ of the noise $\bfZ$ must scale linearly in the sensitivity of $\bfC$ to achieve a target DP guarantee, so our final measures of noise account for this: %
{\small %
\begin{align}
\maxlossop(\bfB, \bfC) &\coloneqq \maxerrop(\bfB) \cdot \sens_\Pi(\bfC) \label{eq:maxloss} \\
\rmslossop(\bfB, \bfC) &\coloneqq \rmseop(\bfB) \cdot \sens_\Pi(\bfC) \label{eq:rmsloss}
\end{align}
}%
\cref{eq:maxloss,eq:rmsloss} measure the distribution of noise to approximate the privacy-utility trade-off as the loss, which do not depend on the noise multiplier $\alpha=\sigma/\sens_\Pi(\bfC)$ that is directly used in accounting for the DP guarantees, as discussed in \citep[Introduction]{mcmahan2024efficient}. For optimized $\bfC$ in matrix factorization mechanisms (e.g., \treeagg, \bandmf, and \blt), specific DP guarantees are achieved by scaling $\alpha$ (and corresponding $\sigma$ in \cref{algo:dpfl}). The total noise on the prefix sum $BZ$ also scales \maxloss and \rmsloss by $\alpha$. Hence, without loss of generality, we use \rmsloss and \maxloss to compare the optimality of different matrix factorization mechanisms, which is equivalent to assuming $\alpha=1$ that corresponds to $(\epsilon=5.3, \delta=10^{-7})$-DP (for example).

Note we deviate from the definitions of \citep[arXiv v3]{mcmahan2024efficient}, in order to distinguish the error introduced by $\bfB$ from the total loss (which is what we optimize and use to compare mechanisms), which incorporates the sensitivity of $\bfC$.

\ifthenelse{\boolean{acl}}{\subsection{Optimizing Multi-participation \blts}}{\subsection{Optimizing \blts for Multiple Participations}} \label{sec:opt_blt}

\ifthenelse{\boolean{acl}}{}{\input{alg_opt_blt}}

For the typical scale of $\nd < 10^5$ in FL systems, rather than deriving closed forms for sensitivity and error as in \citet{mcmahan2024efficient}, we use an alternative approach that is flexible and simple to implement. 
Recall the properties of BLTs discussed in \cref{sec:background_blt}, we parameterize the optimization of the \blt by the pair $(\theta, \hth)$.  \citet[Lem. 5.2]{mcmahan2024efficient} implies given a pair $(\theta, \hth)$, there exist unique $(\omega, \homega)$ such the $\bltm(\theta, \omega)\inv = \bltm(\hth, \homega)$, and we can compute $\omega$ and $\homega$ in time $\calO(\nbuf^2)$; this result is summarized below as \cref{alg:blt_pair}. Thus, given a $(\hth, \theta)$, we can efficiently compute the Toeplitz coefficients of $\bfC$ (using \cref{eq:closedcoefs} applied to $(\theta, \omega)$) and $\bfC\inv$ (applying \cref{eq:closedcoefs} to $(\hth, \homega)$). From the Toeplitz coefficients of $\bfC$ we can then efficiently compute sensitivity using \cref{thm:toepminsepsens}. \rmse can be computed efficiently from the Toepltiz coefficients of $\bfC\inv$ following the approach of \citet[Prop. 3.1] {mckenna2024scaling}, and a simple generalization of this approach applies to to \maxerr as well. For completeness we summarize in the following proposition:

\begin{prop}\label{prop:toeperror}
Let $\bfC = \LtToep(c) \in \R^\dimdim$ be a lower-triangular Toeplitz matrix defined by Toeplitz coefficients $\cc = (c_0, c_1, \dots, c_{\nd-1}) \in \R^\N$. Then $\bfC\inv$ is also a lower-triangular toeplitz matrix; let $\bfC\inv = \LtToep(\hc)$. Then, $\bfB \coloneqq \bfA \bfC\inv = \LtToep(b)$ where $b_i = \sum_{j=0}^{i-1} \hc_i$. Further, we can compute 
{\small \[ 
\maxerrop(\bfB) 
= \sqrt{\sum_{i \in [\nd]} b_i^2}
\]}
and
{\small \[ 
\rmseop(\bfB) 
= \sqrt{ \sum_{i \in [\nd]} (n - i) b_i^2 / \nd}.
\]}

\end{prop}

\ifthenelse{\boolean{acl}}{}{\input{subsec_opt_eff}}

%% file: subsec_blt_sens.tex
\paragraph{Adjacent Data Streams and Privacy Accounting} We assume users (FL clients) participate in training according to a \emph{participation schema} $\Pi \subset \text{Powerset}([\nd])$, where each \emph{participation pattern} $\pi \in \Pi$ (and so $\pi \subseteq [\nd]$) indicates a set of indexes of steps in which a single user might participate. 
Each $\Pi$ results in a adjacency relation $\bfN_\Pi$ on data streams:
two data streams $\obs$ and $\obsp$ are adjacent, that is $(\obs, \obsp) \in \bfN$, if there exists a $\pi \in \Pi$ such that $\obs_t = \obsp_t$ for $t \notin \pi$, and $\norm{\obs_t - \obsp_t}_2 \le \clipnorm$ for $t \in \pi$. 
In FL for user-level DP (~\cref{algo:dpfl}), $\obs_t:=\sum_i \Delta_i^t$ is a sum over per-user model gradients each subject to an L2-norm bound $\clipnorm$, and two streaming datasets are adjacent if one can be formed from the other by ``zeroing out'' all the gradient contributions from any one user following Defn.~1.1 of~\citet{kairouz21practical}.  
Under this adjacent relationship, the DP guarantees of MF mechanism in DP-FTRL can be accounted for the release of $\bfC \bfX + \zeta \bfZ$ according \cref{eq:matrixmech}, computing the sensitivity of $\bfC \bfX$ to calibrate with the Gaussian noise $\bfZ$ of zero mean and $\sigma$ standard deviation~\citep{choquette2023amplified}. 

\paragraph{Multi-participation Sensitivity} We consider \bparticipation, where the distance between any two participations is \emph{at least} b and there are at most $\maxpart$ total participations, formally 
\[
  \Pi_{\minsep, \maxpart} = \left\{\pi \subseteq [n]  \mid \abs{\pi} < \maxpart, \{i, j\} \subseteq \pi, i \ne j \Rightarrow |i - j| \ge \minsep\right\}.
\]
This is motivated not only by the applicability to federated learning (as discussed by \citet{choquette2023amplified}, which also formalized this schema), but also because (implicitly) it is the participation schema under which \treeagg was extended to multiple participations by \citet{kairouz21practical}.

Let $\deltaset := \left\{ \obs - \obsp \mid (\obs, \obsp) \in \bfN \right\}$ represent the set of all possible differences between adjacent $\obs,\obsp$. Then, the L2 sensitivity of $\bfC$ under $\bfN$ is given by
\begin{equation}\label{eq:sensgeneral}
\sens_{\bfN}(\bfC) = \sup_{(\obs, \obsp) \in \bfN} \| \bfC \obs - \bfC \obsp \|_F
 = \sup_{\contrib \in \deltaset} \| \bfC\contrib \|_F. 
\end{equation}

In this work, we only consider $\bfC \ge 0$ (elementwise), and so the supremum over $\contrib$ in \cref{eq:sensgeneral} will always be achieved by some $\contrib \ge 0$ (observe each non-zero entry $\contrib_i \in \R^\mdim$ can be chosen arbitrarily from the unit ball of radius $\clipnorm$). The non-negativity also implies $\bfC\tp \bfC \ge 0$, and hence following Corollary~2.1 of \citet{choquette22multiepoch}, we have
\begin{equation} \label{eq:sens}
  \sens_{\bfN_\Pi}(\bfC) = \clipnorm \max_{\pi \in \Pi} \norm{\bfC \ind(\pi)}_2 \qquad \text{when} \qquad \bfC \ge 0. 
\end{equation}
where $\ind(\pi) \in \{0, 1\}^\nd$ is given by $\ind(\pi)_i = 1$ if $i \in \pi$ and $0$ otherwise. Note that $\clipnorm$ simply introduces a linear scaling, and so we can take $\clipnorm=1$ w.l.o.g. both when optimizing mechanisms and when computing \maxloss and \rmsloss.


%% file: subsec_sens_lower_bound.tex
\paragraph{A Sensitivity Lower Bound} Inspired by \cref{thm:toepminsepsens}, we state a sensitivity lower bound for general matrix in \cref{rem:senslb}. 
An overly optimistic (instead of commonly worst-case) DP guarantees can be computed for MF mechanism with sensitivity in \cref{rem:senslb}. We \emph{only} use \cref{rem:senslb} for the privacy accounting of baseline binary tree mechanisms in simulation experiments in \cref{sec:exp} as the dynamic programming accounting in \citep{kairouz21practical} is computationally expensive. 
In practice we find the lower-bound of \cref{rem:senslb} is tight for the binary tree matrices we consider; proving this is an interesting open problem.
\begin{rem}\label{rem:senslb}
Letting $\pistar \in \Pi$ as in \cref{eq:badpi}, for any mechanism $\bfC$,
\begin{equation}\label{eq:senslb}
\sens_\Pi(\bfC) \geq \norm{\bfC u(\pistar)}_2 
\end{equation}
is a lower-bound on sensitivity (the actual sensitivity might be higher). While \citet{kairouz21practical} introduced a dynamic program for computing binary-tree sensitivity, it requires some work to extend it to the tree completion trick, and in practice it is expensive to compute. Hence, when evaluating \treeagg approaches, for simplicity we use the lower bound of \cref{eq:senslb}, which can be computed immediately for the tree-completion matrices $\bfC$ used when $\nd$ is not a power of $2$.
\end{rem}

%% file: alg_opt_blt.tex
\begin{algorithm}[htbp]
\caption{Differentiable Loss for BLTs} \label{alg:blt_loss}
\begin{algorithmic}
\STATE \textbf{Inputs:} 
\STATE \myind Pair of buffer-decay parameters $(\theta, \hth)$ with $\nbuf$ buffers each ($\theta, \hth \in [0, 1]^\nbuf$).
\STATE \myind num rounds $\nd$, min-separation $\minsep$, max participations $\maxpart$
\STATE \myind Penalty strength $\lambda$, set to zero for loss calculation, or $\lambda = 10^{-7}$ to stabilize optimization
\STATE \textbf{Outputs:}

\STATE \myind Either $\maxlossop(\bfA \bfC\inv, \bfC)$ or $\rmslossop(\bfA \bfC\inv, \bfC)$.
\STATE
\STATEComment Use \cref{alg:blt_pair} to calculate the unique $\omega$ and $\homega$ such that $\bfC = \bltm(\theta, \omega)$ and $\bfC\inv = \bltm(\hth, \homega)$
\STATE $\omega = \mathtt{calc\_output\_scale}(\theta, \hth)$
\STATE $\homega = \mathtt{calc\_output\_scale}(\hth, \theta)$
\STATE
\STATEComment Compute $\var{sens} = \norm{\bfC \pistar}_2$ where $\bfC = \LtToep(c)$ 
\STATE Compute $c \in \R^n$ where $c_0 = 1$ and $
c_i = \sum_{j \in [\nbuf]} \omega_j \theta_j^{i-1}$ for $i \in \{1, \dots, \nd\}.$
\STATE $\bar{c} = 0 \in \R^\nd$  \Comment Holds the sum of columns of $\bfC$
\FOR{$i \in [\maxpart]$}
  \STATE $\bar{c}{[b\cdot i:]}\  +\!\!=\  c{[0 : n - b\cdot i]}$  \Comment \texttt{numpy}-like semantics
\ENDFOR
\STATE $\var{sens} = \norm{\bar{c}}_2$ \Comment{Because $\bar{c} = \bfC \pistar$.}
\STATE
\STATEComment Compute $\text{Errror}(\bfA \bfC\inv)$ where $\bfC\inv = \LtToep(\hc)$.
\STATE Compute $\hc \in \R^n$ where $\hc_0 = 1$ and $
\hc_i = \sum_{j \in [\nbuf]} \homega_j \hth_j^{i-1}$ for $i \in \{1, \dots, \nd\}.$
\STATE Compute $b \in \R^n$ by $b_i = \sum_{j=0}^{i} \hc_i$  \Comment So $\bfB = \bfA\bfC\inv = \LtToep(b)$.
\STATE $\var{err} = \begin{cases}
\sqrt{\sum_{i \in [\nd]} b_i^2} & \text{for \maxerr} \\
\sqrt{ \sum_{i \in [\nd]} (n - i) b_i^2 / \nd} & \text{for \rmse}.
\end{cases}$

\STATE
\STATEComment Log-barrier penalties to keep $\theta > 0$, $\theta < 1$, and $\omega > 0$ for numerical stability when optimizing
\STATE $\var{penalty} = \lambda (-\log(\theta) -\log(1 - \theta) -\log(\omega))$

\STATE 
\STATE \textbf{Return} $\var{loss} = \var{err} \cdot \var{sens} + \var{penalty}$
\end{algorithmic}
\end{algorithm}

\begin{algorithm}[htbp]
\caption{\texttt{calc\_output\_scale}  (Lemma 5.2 of \citet{mcmahan2024efficient})} \label{alg:blt_pair}
\begin{algorithmic}
\STATE \textbf{Input:} 
\STATE \myind Pair of buffer-decay parameters $(\theta, \hth)$ with $\nbuf$ buffers each ($\theta, \hth \in [0, 1]^\nbuf$).

\STATE \textbf{Output:}
\STATE \myind The unique $\omega$ s.t. $\bfC = \bltm(\theta, \omega)$ has a \blt inverse with buffer-decay $\hth$ ($\bfC\inv = \bltm(\hth, \cdot)$).
\STATE
\STATEComment Note that in \citep{mcmahan2024efficient}, $\bfC$ was parameterized by $\hth$ and $\bfC\inv$ was parameterized by $\theta$.
\STATE $p(x) = \prod_{i \in [\nbuf]} (1 - \hth_i x)$
\STATE $q(x) = \prod_{i \in [\nbuf]} (1 - \theta_i x)$
\STATE $f(x) = (p(x) - q(x)) / x$  \Comment Polynomial division gives $f$, a polynomial of degree $\nbuf-1$
\STATE $z = \prod_{i \in [\nbuf]} -\theta_i$ 
\STATE $w_i = \left(\prod_{j \ne i} (\theta_i\inv - \theta_j\inv)\right)\inv$ for $i \in [\nbuf]$
\STATE Define $\omega$ by $\omega_i = f(\theta_i\inv)\frac{-\theta_i w_i}{z}$ for $i \in [\nbuf]$
\STATE \textbf{Return} $\omega$

\end{algorithmic}
\end{algorithm}

%% file: subsec_opt_eff.tex
\begin{table*}[htbp]
  \begin{varwidth}[b]{0.54\linewidth}
    \centering
    \begin{tabular}{rrrr}
    \toprule
    $n$ & brute force & via \cref{alg:blt_pair} & speedup \\
    \midrule
    2000 & 0.021 & 3.8e-5 & 550$\times$ \\
    20000 & 0.258 & 3.6e-5 & 7176$\times$ \\
    200000 & 4.772 & 4.5e-5 & 104884$\times$ \\
    \bottomrule
    \end{tabular}
    \caption{Seconds to compute $n$ Toeplitz coefficients of $\bfC\inv$. JAX just-in-time (JIT) compilation is \emph{not} included for either approach.}    
    \label{tab:c_inv_coef}
  \end{varwidth} %
  \hfill
  \begin{minipage}[t]{0.44\linewidth}
    \centering
    \includegraphics{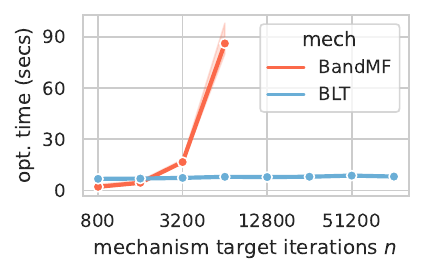} %
    \captionof{figure}{Total wall clock optimization time including JIT compilation on a V100 GPU for \bandmf and \blts, fixing $\minsep=400$ and varying $\nd$. The average over 3 runs is reported.}
    \label{fig:faster}
  \end{minipage}%
\end{table*}

Combining these elements gives us an efficient and differentiable algorithm for computing \maxloss and \rmsloss. Complete pseudo-code for the differentiable loss calculation is given in \cref{alg:blt_loss}. Following \citet{mcmahan2024efficient}, we use auto differentiation and L-BFGS optimizer in JAX~\citep{jax2018github} to optimize $(\theta, \hth)$ for the BLT-DP-FTRL algorithm, and then extract $\bltm(\theta, \omega)$ for noise generation. Similar to \citep{mcmahan2024efficient}, we introduce log-barrier penalties w/ strength $10^{-7}$ to keep $\omega > 0$, $\theta > 0$ and $\theta < 1$ (which is necessary to ensure the Toeplitz coefficients of $\bfC$ are decreasing to satisfy \cref{thm:toepminsepsens}).  
For high precision optimization, we use double precision in JAX on CPUs and GPUs. We observe that increasing buffer size $d$ does not necessarily reduce the loss due to numerical stability and optimization challenges, and different BLT parameter $(\theta, \omega)$ may be achieved in different optimization runs. We also highlight that the different BLT parameters $(\theta, \omega)$ can generate similar Toeplitz coefficients for $\bfC$, which suggests a smaller $d$ might help mitigate the optimization challenge from overparametrization.

The primary motivation for utilizing the $(\theta, \hth)$ parameterization in \cref{alg:blt_loss} is computational efficiency. \cref{tab:c_inv_coef} compares the time to compute $n$ Toepltiz coefficients for $\bfC\inv$ given either $\bltm(\theta, \omega)$ or given $(\theta, \hth$). In the first caes (``brute force''), we construct the Toeplitz coefficients of $\bfC$ using \cref{eq:closedcoefs}, and then solve a linear system (using \texttt{jax.lax.scan} and the property that the inverse of a lower triangular Toeplitz matrix is also lower triangular Toeplitz) to compute the coefficients of $\bfC\inv$). In the second case, we use \cref{alg:blt_pair} to compute $(\omega, \homega)$, and then compute the Toeplitz coefficients by applying \cref{eq:closedcoefs} to $\bltm(\hth, \homega)$. The comparison uses a V100 GPU and a compiled \texttt{jax} implementation, and is repeated many times with an average is given. The second approach can be fully vectorized, and is orders of magnitude faster. This is critical because this is the primary computational step in computing the \rmsloss or \maxloss and occurs in the inner loop of the mechanism optimization procedure: the net result is mechanism optimization is substantially faster than for \bandmf, and scales to much larger $\nd$, see \cref{fig:faster}. \cref{alg:blt_loss} does incur more \texttt{jax} just-in-time (JIT) compilation overhead compared to \bandmf optimization, which accounts \blt optimization being slightly slower for small $\nd$.

%% file: sec_exp.tex
\section{Simulation Experiments} \label{sec:exp}

\input{tab_sonwp}

We run simulation experiments before applying our BLT-DP-FTRL algorithm in \cref{sec:background_blt} to train production LMs. The BLT parameters $(\theta, \omega)$ are optimized with our multi-participation approach in \cref{sec:blt_mp}. We present private-utility trade-off on StackOverflow benchmark dataset in \cref{sec:sonwp}, and \maxloss, \rmsloss across a range of scenarios in \cref{sec:rmse_exp}. We compare \blts to both flexible \treeagg~\citep{kairouz21practical} and state-of-the-art \bandmf~\citep{choquette2023amplified} (see \cref{sec:background} for more discussion). 
In the simulation experiments, we are \emph{maximally generous} in evaluating \treeagg mechanisms, considering the memory cost to be $\ceil{\log_2 \nd}$, while calculating \rmse and \maxerr using the optimal \tafull~\citep{denisov22matfact} without memory-efficient implementation, and use the lower bound of \cref{rem:senslb} to account for overly optimistic privacy-utility trade-off. Thus, in all cases we over-estimates the true performance of the binary tree, but nevertheless we show \blts have superior performance in terms of both error and memory.

\begin{figure}[htb]
    \centering
    \includegraphics[width=0.9\linewidth]{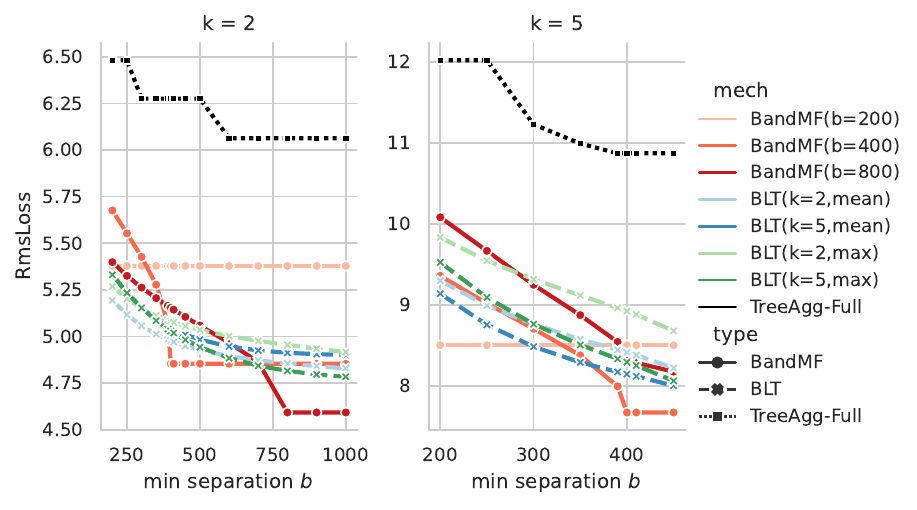}
    \caption{\small Comparison of mechanisms in terms of prefix-sum root-mean-squared error at a fixed privacy level. \blts were optimized for either \rmsloss (``mean'') or \maxloss (``max''), and for either $\maxpart=2$ or $\maxpart=5$ participations at min-separation $\minsep=400$. BandMF matrices were optimized for $\minsep \in \{200, 400, 800\}$ (the \bandmf optimization does not depend on $\maxpart$, and previous work optimizes for \rmsloss). 
    We also include \tafull using the optimal (``full Honaker'') decoding (for which a memory-efficient noise generation algorithm is unknown). 
    \textbf{We observe that all the \blts perform competitively with \bandmf, and can outperform \bandmf} when the min-separation differs significantly from the number of bands. For example, with $k=2$ participations (left panel) our BLT($k=2$,mean) (light blue) BLT outperforms BandMF($b=400$) when min separation is less than 390 or greater than 700. 
    }
    \vspace{-0.4cm}
    \label{fig:rmse}
\end{figure}

\subsection{StackOverflow NWP Benchmark} \label{sec:sonwp}

We follow \citet{choquette2023amplified} for StackOverflow next word prediction (NWP) experiments, including all hyperparameter tuning, and vary only the DP mechanism to compare \bandmf, \tafull, and \blts. Results are summarized in \cref{tab:sonwp}. \bandmf still achieves the highest performance, as in this scenario we train for a fixed known number of rounds, with an exactly known max participations $\maxpart=6$ and min-separation $\minsep=342$. \tafull and \blts optimized for only $\maxpart=1$ participation~\citep{mcmahan2024efficient} are noticeably worse, but our multi-participation-optimized \blts are very competitive with \bandmf with only 2 or 3 buffers (with a $171\times$ and $114\times$ reduction in runtime memory overhead). 
\ifthenelse{\boolean{acl}}{}{We expect the negligibly worse performance of larger numbers of buffers may be due to the impact of regularization during the optimization of the \blt parameters.}
In the relatively large signal-to-noise ratio regime ($\epsilon=8$), \ourblt achieves comparable or even better learning accuracy though the \rmsloss (\maxloss) is slightly worse.

\subsection{\rmsloss and \maxloss Experiments} \label{sec:rmse_exp}

\paragraph{Comparing BLT to \tafull and \bandmf} We further show that \blt is better than {\small \tafull}, and more flexible than \bandmf by computing \rmsloss (\maxloss) in a wide range of scenarios. Because the \bandmf mechanisms are optimized for \rmsloss, we compare on this metric in \cref{fig:rmse}. However, in both our StackOverflow and Gboard experiments described subsequently, we deploy BLT mechanisms optimized for \maxerr following \citep{mcmahan2024efficient}. For completeness, we provide \cref{fig:maxloss} in \cref{app:plots} that compares the mechanisms on \maxloss. 
\ifthenelse{\boolean{acl}}{}{The BLTs directly optimized for the metric perform better than optimizing for a different metric.} 
We observe that \emph{all the \blts perform competitively with \bandmf, and can outperform \bandmf} when the min-sep differs significantly from the number of bands. For example, in \cref{fig:rmse}, with $k=2$ participations (left panel) our BLT($k=2$,mean) (light blue) BLT outperforms BandMF($b=400$) when min separation is less than 390 or greater than 700. 

\ifthenelse{\boolean{acl}}{We provide more results on the robustness of \blts to min-sep $b$, number of rounds $n$, and comparing with \bandtoep in \cref{app:mse}.  }{\input{subsec_mse}}

%% file: tab_sonwp.tex
\begin{table}[tbh]
    \centering
\begin{small}
\begin{tabular}{lrrrr}
\toprule
& \multicolumn{2}{c}{Test Accuracy } & RMS & Max \\
Mechanism &  $\epsilon = 2$ &  $\epsilon = 8$ & Loss & Loss\\

\midrule
\bandmf(band=342)\ifthenelse{\boolean{acl}}{}{~\citep{choquette2023amplified}} & 23.21 & 24.86 & 10.21 & 8.60 \\
\tafull\ifthenelse{\boolean{acl}}{}{~\citep{denisov22matfact}} & 22.54 & 24.47 & 14.98 & 12.47 \\
\blt(nbuf=2,$k$=1)\ifthenelse{\boolean{acl}}{}{~\citep{mcmahan2024efficient}} & 22.37 & 24.64 & 11.80 & 11.15 \\
\blt(nbuf=5,$k$=1)\ifthenelse{\boolean{acl}}{}{~\citep{mcmahan2024efficient}} & 22.40 & 24.63 & 11.40 & 10.87 \\
\blt*(nbuf=2,$k$=6) & 23.09 & 24.83 & 10.81 & 9.34 \\
\blt*(nbuf=3,$k$=6) & 23.13 & 24.87 & 10.79 & 9.33 \\
\blt*(nbuf=4,$k$=6) & 23.13 & 24.83 & 10.79 & 9.33 \\
\blt*(nbuf=5,$k$=6) & 23.07 & 24.84 & 10.79 & 9.33 \\
\bottomrule
\end{tabular}
\end{small}
    \caption{\small Comparing mechanisms in terms of test-set accuracy on the StackOverflow NWP task. All runs are based on $\nd=2052$ rounds of training with $\maxpart=6$ participations and min-sep $\minsep=342$. \blts are optimized for \maxloss. Results are visualized in \cref{fig:sonwp} in \cref{app:plots}.}
    \label{tab:sonwp}
    \vspace{-0.4cm}
\end{table}

%% file: subsec_mse.tex
\begin{figure}[H]
    \centering
    \includegraphics[width=\linewidth]{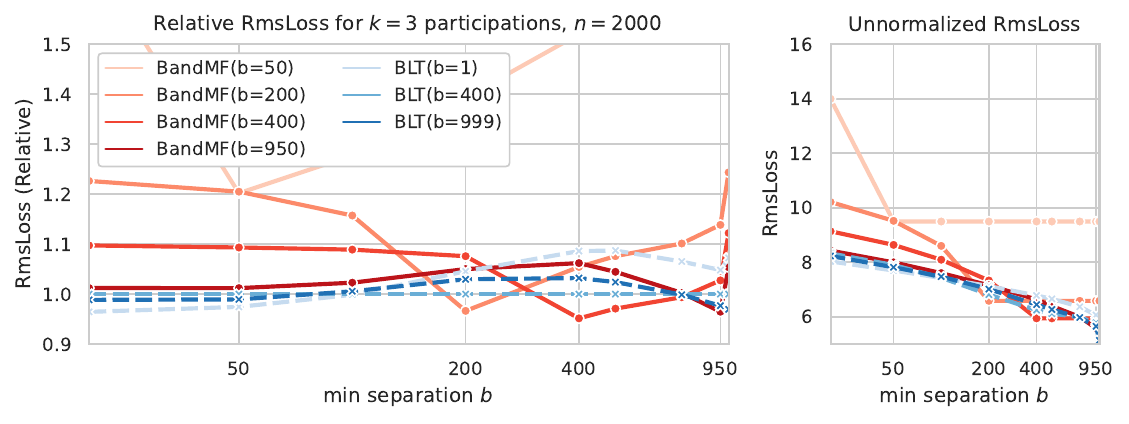}
    \caption{Comparison of \bandmf and \blts with $\nd = 2000$ and $k=3$ participations, varying the min-separation $\minsep$. The \blts were optimized for \rmsloss. The $x$-axis is shown on a log-scale. The left panel gives loss relative to the $\bltm(b=400)$ mechanism, while the right panel gives the same data on an unnormalized $y$-axis scale.}
    \label{fig:rmse_k3}
\end{figure}

\begin{figure}[H]
    \centering
    \includegraphics[width=\linewidth]{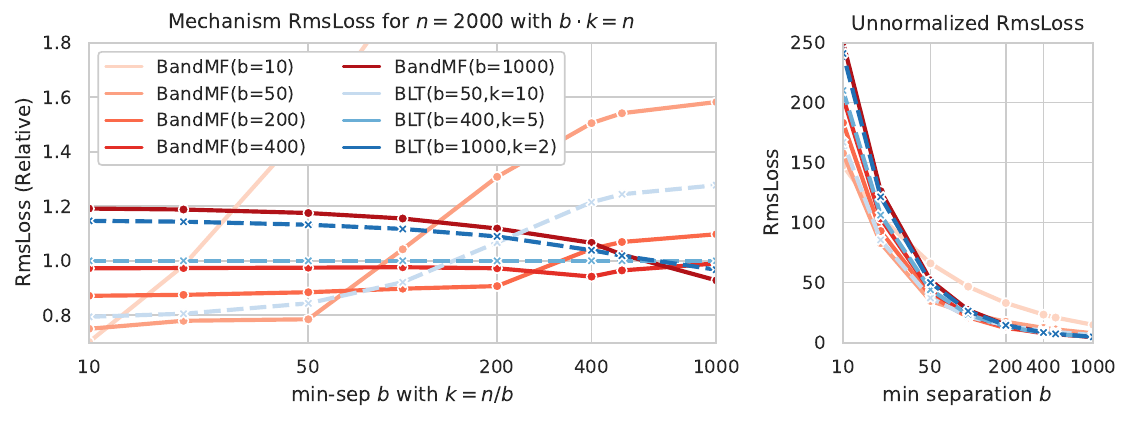}
    \caption{Comparison of \bandmf and \blts with $\nd = 2000$, varying $\minsep$ such that $\maxpart = \nd / \minsep$ is an integer. The \blts were optimized for \rmsloss. The $x$-axis is shown on a log-scale. The left panel gives loss relative to the $\bltm(b=400)$ mechanism, while the right panel gives the same data on an unnormalized $y$-axis scale.}
    \label{fig:rmse_nkb}
\end{figure}

\paragraph{Robustness to Min-sep $b$} \cref{fig:rmse_k3} compares BLT to the strong baseline \bandmf, fixing $\nd = 2000$ and $k=3$ participations and varying the min-separation $\minsep$. We make three primary observations: 
\begin{enumerate*}[label=\color{purple}(\arabic*)]
\item 
The optimization of the \bandmf mechanisms implicitly assumes $b \sim \hat b$, and
as expected, near this regime \bandmf in fact (slightly) outperforms the \blts. 
\item However, when $b \ll \nd / \maxpart$,  the \blts perform better. 
\item Interestingly, \bandmf performs significantly worse than the \blts at $b=999$. In this case, the only participation patterns that actually have 3 participations are e.g. $\{0, 999, 1998\}$, $\{0, 1000, 1999\}$ --- importantly, only the columns $\{0, 1, 999, 1000, 1998, 1999\}$ can ever occur in a 3-participation pattern. Because the columns of $\bfC$ for \bandmf all have the same column norm, this fact cannot be exploited. However, because of their Toeplitz structure, columns 1998 and 1999 have smaller norms than other columns, and that is beneficial in this situation. 
\end{enumerate*}

The setting where $\maxpart$ is fixed and we vary $\minsep$ includes situations that should generally be avoided in practice. For example, if we had $\maxpart=3$, $\minsep=10$, and $\nd=2000$, this indicates we had enough data we should have been able to achieve $\minsep = \nd/\maxpart \approx 667$, and so $\minsep=10$ indicates a highly suboptimal data ordering.  Similarly, if we had $\maxpart=3$, $\minsep=999$, and $\nd=2000$, then we would have been better off stopping at $n=1998$, which would have ensured only $\maxpart=2$ participations and significantly decreased sensitivity (at presumably a small cost in learning performance compared to training for $\nd=2000$ iterations).

\cref{fig:rmse_nkb} shows the contrasting scenario (indicating an essentially optimal data ordering, general not possible in federated learning) that occurs when we fix $\nd = 2000$, and choose $\minsep$ that exactly divide $\nd$ so that we can take $\maxpart = \nd / \minsep$ exactly. \cref{fig:rmse_nkb} considers the worst-case max participation $k$ for given min-sep $b$ and total rounds $n$, and achieves generally larger \rmsloss. When $b \leq \hat b$, \bandmf slightly outperforms \blts, but \bandmf degrade more rapidly for $b \leq \hat b$. In general, the curves of \blts are more smoother across different min-sep $b$ in both \cref{fig:rmse_k3} and \cref{fig:rmse_nkb}.

\begin{figure}[htb]
    \centering
    \includegraphics[width=\linewidth]{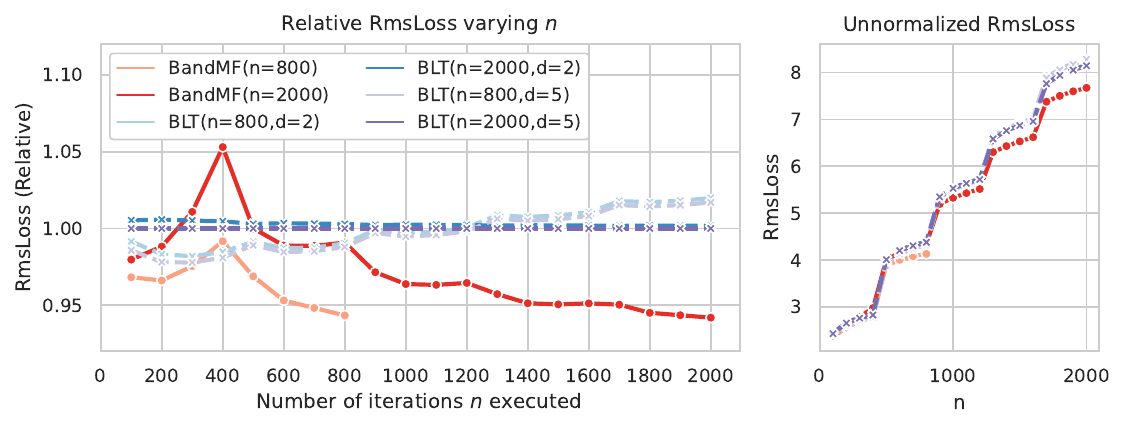}
    \caption{Comparison of \bandmf and \blts optimized for $\nd=800$ and $\nd=2000$, and evaluated for $\nd$ up to the optimization target (for $\bandmf$) and over $[0, 2000]$ for the \blts. All mechanisms were optimized for min-separation (bands) $\minsep=400$. \blts with $\nbuf=2$ and $\nbuf=5$ perform almost equivalently; $\nbuf=1$ (not shown),is not sufficient with relative \rmsloss $> 1.07$.}
    \label{fig:rmse_vary_n}
\end{figure}

\paragraph{Robustness to Total Rounds $n$} \cref{fig:rmse_vary_n} considers varying the number of steps of the mechanism actually executed for mechanisms optimzied for different $\nd$. \bandmf mechanisms can only be used up to the $\nd$ they are optimized for, but \blts naturally extend to any $\nd$. This figure demonstrates that again \blts are not particularly sensitive to the $\nd$ for which they are optimized. For this figure, the maximum number of participations is chosen to be the largest allowed given $\nd$ and $\minsep=400$ (i.e., $k=n/d$), leading to the stairstep behavior of the unnormalized \rmsloss in \cref{fig:rmse_vary_n} (Right). \bandmf optimizing for large $n$ performs well when the actual number of iterations executed is small, but optimizing for large $n$ encounters nontrivial as discussed in \cref{tab:mechanism_summary} and \cref{fig:faster}. Finally, these results show that only $\nbuf=2$ buffers is sufficient for good performance, or a $200\times$ memory savings compared to \bandmf with $\minsep=400$ bands.


\begin{figure}[htb]
    \centering
    \includegraphics[width=\linewidth]{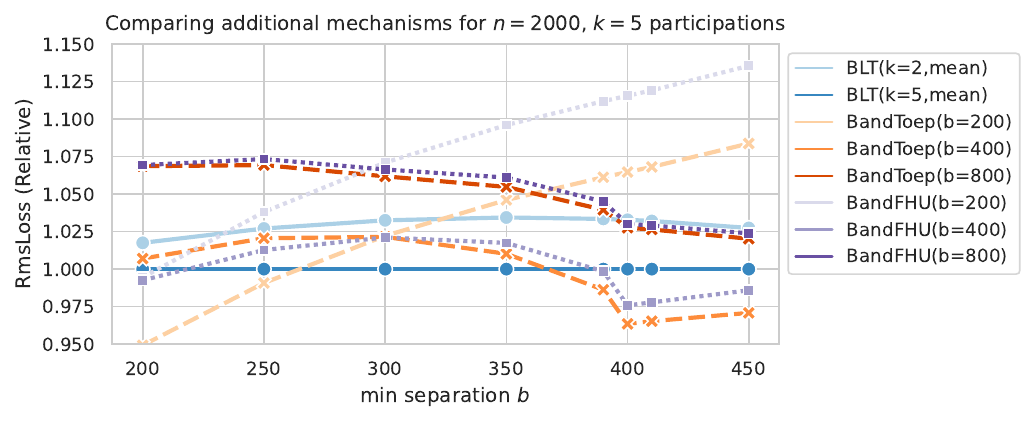}
    \caption{Comparison of mechanisms for $\nd=2000$ and $\maxpart=5$ participations, optimized for min-separation $\minsep = 400$ and compared for different actual values of min-separation. The setting is comparable to that of \cref{fig:rmse}, so only relative lines are given. 
    }
    \label{fig:other_mechs}
\end{figure}

\paragraph{Comparing \blt to \bandtoep and \bandfhu} Finally, we compare BLT-DP-FTRL to several other more recent mechanisms:
\begin{itemize}
    \item \bandtoep~\citep{mckenna2024scaling}, which optimizes banded Toeplitz matrices $\bfC$ for \bparticipation under \rmsloss. The primary advantage of this mechanism compared to \bandmf is that \bandtoep matrices can be optimized for much larger $\nd$. However, the runtime is the same as \bandmf, and the optimization is still slower than the optimization of \blts.
    \item \bandfhu~\citep{kalinin2024banded}, which uses prefixes of the optimal-for-single-participation \maxloss coefficients of \citet{fichtenberger2022constant} to form banded Toeplitz matrices. These will likely be worse than \bandtoep (which is specifically optimized for multiple participations), but require no mechanism optimization.
\end{itemize}
\cref{fig:other_mechs} shows that \blts are comparable or better to both of these approaches.

%% file: sec_gboard.tex
\section{Production LMs for Mobile Keyboard} \label{sec:gboard}

\input{tab_gboard}

\paragraph{Production Setting} Our BLT-DP-FTRL algorithm is flexible in min-sep (shown in \cref{sec:rmse_exp}), achieves competitive privacy-utility performance for relatively large signal-to-noise ratio (shown in \cref{sec:sonwp}), and saves computation and memory cost (shown in \cref{tab:mechanism_summary}), which motivates the usage in production FL systems. 
\ifthenelse{\boolean{acl}}{ We follow~\citep{xu2023federated} for the production setting, and provide additional details including the configuration for baselines in \cref{app:prod_config}.
}{
\input{subsec_gboard_config}
}
\input{fig_examp_curve}

\ifthenelse{\boolean{acl}}{\textbf}{\paragraph}{Main Results} We summarize the privacy and utility results for es-ES, id-ID, pt-BR, and pt-PT LMs in \cref{tab:gboard}, and show the privacy-utility curves for training pt-PT in \cref{fig:example_curve}. We provide additional curves for other LMs in \cref{fig:acc_curve,fig:acc_curve2,fig:priv_curve} in\cref{app:plots}, and discuss the observations here.
\begin{enumerate*}[label=\color{purple}(\arabic*)]
    \item \emph{Most of the models achieve DP guarantee of $\epsilon<10$ with exception of $\epsilon \sim 10$ for pt-PT due to the challenge of small population; the pt-BR model trained with \blt-16.1 achieves $\epsilon<1$ at round 2000.} DP guarantees of $\epsilon<10$ is commonly used for machine learning, and $\epsilon<1$ are considered strong guarantees~\citep{ponomareva2023dpfy}. To achieve single-digit DP guarantees in practice without sacrificing the utility, the production LMs are trained with large number of clients per round (known as report goal in FL systems). We use typical report goal 6.5K~\citep{xu2023federated} for es-ES, id-ID and pt-BR, and use a smaller report goal 3K for pt-PT with a smaller population. We additionally run BLT-DP-FTRL with larger report goal and linearly scale up the noise multiplier to keep the signal-to-noise ratio for utility: \blt-16.1 with report goal 12K for pt-BR and \blt-5.5 with report goal 6K for pt-PT. The resulting min-seps in pt-BR and pt-PT almost halved when the report goals are increased. We use noise multiplier 7 for \treeagg, which is determined by StackOverflow simulation experiments~\citep{xu2023federated}. \ifthenelse{\boolean{acl}}{}{The noise multiplier of \bandmf is computed so that the RMSE is similar to \treeagg with noise multiplier 7~\citep{choquette2023amplified} and will achieve stronger DP guarantees. As \blt achieves comparable privacy-utility trade-off as \bandmf in the high single-to-noise ratio regime ($\epsilon=8$) in simulation experiment (\cref{tab:sonwp} ), the noise multiplier of \blt is computed so that \blt achieves similar DP guarantee as \bandmf for the desired round.}
    \item \emph{\blt achieves better privacy-utility trade-off compared to \treeagg and \bandmf.} \blt achieves comparable and slightly better NWP accuracy for the models in \cref{tab:gboard} and  \cref{fig:example_curve}, and stronger DP guarantees. The model performance are further verified by A/B test in the production application comparing \blt models to baseline models for WMR and WPM, where we target on improved or neutral utilities. The advantage of \blt in privacy-utility trade-off is clearly demonstrated in \cref{fig:priv_curve}, and \blt is better than not only \treeagg, but also \bandmf across the production LM training. 
    The practical min-sep can be quite different from the estimated min-seps for optimizing \bandmf and \blt matrices, \ifthenelse{\boolean{acl}}{e.g., $\sim$300 compared to 400 for es-ES, and 2000+ compared to 1000 for pt-BR}{$\sim$90 compared to 100 for pt-PT,  $\sim$300 compared to 400 for es-ES, $\sim$440 compared to 400 for id-ID, and 2000+ compared to 1000 for pt-BR}. As \blt is more flexible on min-sep estimation, the challenge of reliably estimating min-sep resulting in \blt achieving even stronger privacy-utility trade-offs than \bandmf in the production LMs training. 
\end{enumerate*}

\ifthenelse{\boolean{acl}}{ 
\textbf{Extrapolation} We extrapolate the results for production setting by assuming linearly increase report goal and noise multiplier, and changing min-sep will not change the utility, and hence we can study the effect on DP without actually training the model. We provide results and detailed discussion in \cref{app:extrap}, which further demonstrate the advantages of BLT-DP-FTRL. 
}{
\input{subsec_gboard_extrap}
}

%% file: tab_gboard.tex
\begin{table*}[thb]
    \centering
    \small
    \begin{tabular}{c|c|c|c|c|c|c|c}
         \hline
         \multirow{2}{*}{LM} & \multirow{2}{*}{Rnds} & \multicolumn{3}{c|}{Utility} & \multicolumn{3}{c}{Privacy} \\
         \cline{3-8}
           & & NWP(\%) & WMR(-\%) & WPM(+\%)   & Mech-$\sigma$/MinS/MaxP & zCDP & DP-$ \epsilon$ \\
         \hline
         \multirow{2}{*}{es-ES} & \multirow{2}{*}{1280} & 14.07 $\pm$ 0.06 &  - & - 
            & \bandmf-1.411 / 296 / 4  & 0.29 & 4.82  \\
            & &  13.98 $\pm$ 0.11 & 0.38 & 0.13 & \blt-7.379 / 300 / 4 & 0.16 & 3.46 \\
         \hline
          \multirow{2}{*}{id-ID} & \multirow{2}{*}{2350} & 5.80 $\pm$ 0.10 &  -  & - 
            & \treeagg-7 / 437 / 5  & 0.94 & 9.29  \\
            & & 5.87 $\pm$ 0.04  & 0.09 & 0.07 & \blt-7.379 / 447 / 5 & 0.20 & 3.93 \\
         \hline
         \multirow{3}{*}{pt-BR} & \multirow{3}{*}{2000} & 13.77 $\pm$ 0.36 &  - & - 
            & \bandmf-4.523 / 2001 / 1  & 2.45e-2 & 1.32  \\
            & & 13.86 $\pm$ 0.25 & {\color{gray} -0.04} & 0.04 & \blt-8.681 / 2001 / 1 & 2.23e-2 & 1.25 \\
            & & 13.96 $\pm$ 0.18 & {\color{gray}-0.13} & 0.18 & \blt-16.1 / 1181 / 2 & 1.40e-2 & 0.98 \\
         \hline
         \multirow{3}{*}{pt-PT}  & 430 & 13.58 $\pm$ 0.06 & - & -  
         & \treeagg-7 / 91 / 4 & 1.33 & 11.27 \\
         & 430 & 13.76 $\pm$ 0.11 & 0.38 & {\color{gray} -0.24} & \blt-3.12 / 92 / 4 & 1.11 & 10.19\\
         & 320 & 13.66 $\pm$ 0.04 & 0.07 & {\color{gray} -0.12} & \blt-5.5 / 49 / 6 & 0.75 & 8.19\\
         \hline
    \end{tabular}
    \caption{\small Privacy and utility of production LMs. Utility are measured by NWP accuracy averaged between $r\pm50$ rounds for round $r$ ($r\pm10$ for pt-PT), and the relative WMR decrease and WPM increase in A/B test; privacy shows the key parameters and corresponding DP guarantees, and smaller DP guarantees represent stronger protection; DP-$\epsilon$ is accounted for small $\delta=10^{-10}$; estimated population sizes are es-ES (4.21M), id-ID (8.9M), pt-BR (16.6M), and pt-PT (0.83M). We run additional experiments on pt-BR and pt-PT with larger noise multipliers linearly scales with larger report goal for the \blt mechanism. }
    \vspace{-0.4cm}
    \label{tab:gboard}
\end{table*}


%% file: subsec_gboard_config.tex
Following \citet{hard2018gboard,xu2023federated}, we train one-layer LSTM LMs of $\sim6.4M$ parameters for mobile keyboard applications. These LMs are deployed on device to predict words during decoding time to facilitate user typing. 
We use next word prediction (NWP) accuracy on a hold-out set of devices to track the training progress, and also conduct A/B test in production, where following two metrics are reported:
\begin{enumerate*}[label=\color{purple}(\arabic*)]
    \item Word Modified Ratio (WMR), the ratio of words being modified during typing or after committed; improvement is shown by reduction;
    \item Word Per Minute (WPM): the number of committed words per minute.
\end{enumerate*}
LMs are trained for different language-locale combination in different populations. We study Spanish in Spain (es-ES), Indonesian in Indonesia (id-ID), Portuguese in Brazil (pt-BR) and Portuguese in Portugal (pt-PT). LMs are pre-trained on public multilingual C4 dataset~\citep{xue2020mt5} before private training with FL and DP. 

\paragraph{Algorithm Setting} We compare our BLT-DP-FTRL algorithm with \treeagg~\citep{kairouz21practical,xu2023federated} and \bandmf~\citep{choquette2023amplified} discussed in \cref{sec:background}. As far as we know, these two are the only DP algorithms actively used to train LMs in a production FL system. We follow the system and parameter configurations in~\citep{xu2023federated,choquette2023amplified,gboard_dp_blogpost} for baselines, and compare to \treeagg for pt-PT and id-ID, and \bandmf for es-ES and pt-BR. However, we highlight it is challenging to exactly reproduce the settings, especially for the min-sep parameter. The \bandmf algorithm are optimized for total round $n=2000$, band $\hat b = 400$ for es-ES, and $\hat b = 1000$ for pt-BR. We optimize \blt for total round $n=4000$\footnote{As the target round is usually less than 2000, $n=4000$ for \blt is less favorable compared to $n=2000$ used for \bandmf. \blt is robust to the target round $n$, and achieves stronger results with an inferior $n$.}, estimated min-sep $b=100$ for pt-PT, $b=400$ for es-ES and id-ID, and $b=1000$ for pt-BR. We use \ourblt for multi-participation and estimate the max-par based on $n/b$. For these $n,b$ settings, only $d=4$ buffers can achieve near-optimal loss in optimization, and \ourblt matrices are parameterized by only 8 numbers (these parameters are provided in \cref{app:blt_params}). Though different configures are used for populations with different sizes, the BLT parameters $(\theta, \outp) \in \R^8$ optimized for $b=400$ can achieve competitive results for a wide range of min-seps, which can be a reasonable default for BLT-DP-FTRL. As discussed in \cref{tab:mechanism_summary}, \blt is more memory-efficient than both \treeagg and \bandmf. In the simulation results \cref{tab:sonwp}, \blt is also better than \treeagg for privacy-utility trade-off, and comparable with \bandmf. The results in production further show that the flexibility of \blt makes it easier to use in practice, and achieve better results than both \treeagg and \bandmf.

%% file: fig_examp_curve.tex
\begin{figure}[thb]
\centering
\begin{subfigure}[b]{0.45\linewidth}
\centering
\includegraphics[width=\textwidth]{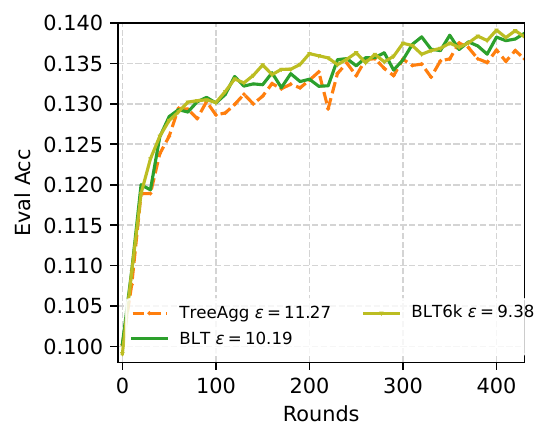}
\caption{\small NWP \ifthenelse{\boolean{acl}}{}{evaluation} accuracy}
\label{fig:example_acc}
\end{subfigure}
\begin{subfigure}[b]{0.44\linewidth}
\centering
\includegraphics[width=\textwidth]{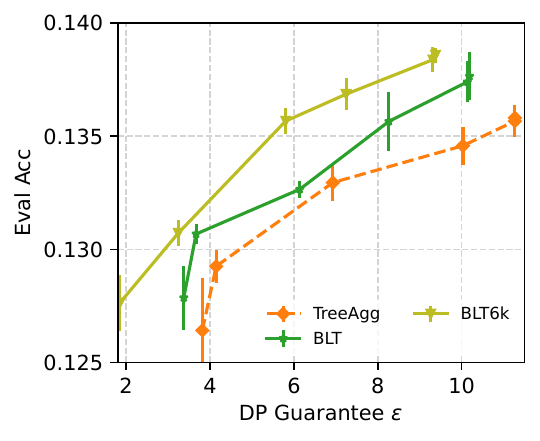}
\caption{\small Privacy-utility trade-off}
\label{fig:example_priv}
\end{subfigure}
\ifthenelse{\boolean{acl}}{
\caption{\small NWP evaluation accuracy and the dervied privacy-utility trade-off curves for training the Portuguese LM in Portugal (pt-PT) with DP-FTRL in a FL system. Additional curves for es-ES, id-ID, and pt-BR are provided in \cref{fig:acc_curve,fig:acc_curve2,fig:priv_curve}. \blts achieve better privacy-utility trade-off.}
}{
\caption{\small NWP evaluation accuracy and privacy-utility trade-off curves for training the production Portuguese LM in Portugal (pt-PT) with DP-FTRL in a FL system. Additional curves for es-ES, id-ID, and pt-BR are provided in \cref{fig:acc_curve,fig:priv_curve}, with NWP accuracy zoomed-in for later stage of training in \cref{fig:acc_curve2}. The privacy-utility trade-off curves are derived from NWP accuracy curves: for each selected round $r$, we compute the mean and standard deviation (shown as vertical bars) for accuracy from the rounds in the range of $r\pm10$, and accounting the DP guarantees. \blts achieve better privacy-utility trade-off.}
}\label{fig:example_curve}
\vspace{-0.4cm}
\end{figure}

%% file: subsec_gboard_extrap.tex
\input{fig_extrap}

\paragraph{Extrapolation} We extrapolate the results for production setting by using a common hypothesis: linearly increase report goal and noise multiplier will not change the utility of the model as the signal-to-noise ratio is maintained. In addition, we assume only changing min-sep will not change the utility because of the signal-to-noise ratio. The hypothesis has been verified in previous work~\citep{kairouz21practical} and the large report goal experiments for pt-BR and pt-PT in \cref{tab:gboard} and \cref{fig:acc_curve}. Hence we can study the effect on DP without actually training the model, similar to \cref{sec:rmse_exp} for simulation.

We vary the report goal, and the min-sep is optimistically estimated by \emph{min-sep=$\lfloor$population-size/report-goal$\rfloor$}, and \emph{max-par=$\lceil$total-rounds/min-sep$\rceil$} is used unless otherwise specified. We discuss the extrapolation results in \cref{fig:extrapolate}, where utility is the same based on the hypothesis.
\begin{enumerate*}[label=\color{purple}(\arabic*)]
\item \blt achieves better DP guarantees because of its robustness to min-seps and total rounds. 
\item We observe that using larger report goal and optimizing for the largest possible min-sep achieves better results than using smaller report goals and larger corresponding min-sep, similar to observation for \treeagg in \citep{xu2023federated}.
\item \cref{fig:pop10m} shows training more rounds does not necessarily increasing DP guarantees when min-sep is large.
\item The gap of \blt and \bandmf is small when min-sep is accurately estimated. In the regime of relatively high signal-to-noise ratio (large noise multiplier for limited computation resources), \blt is competitive in a wide range of different configurations. Hence \blt is easier to use in production FL systems compared to \bandmf, and also saves memory during training.
\end{enumerate*}

\begin{figure}[thb]
\centering
\begin{subfigure}[b]{0.325\linewidth}
\centering
\includegraphics[width=\textwidth]{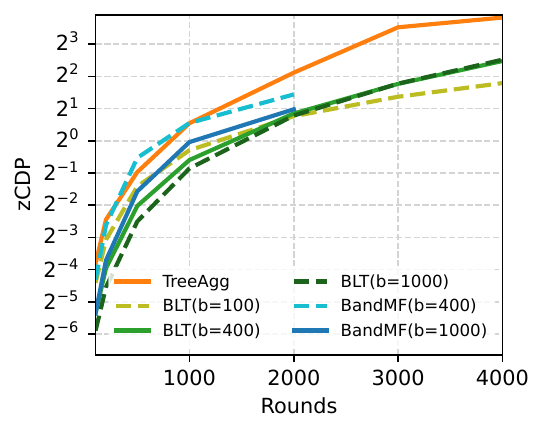}
\caption{\small Min-sep b=100}
\label{fig:varyn_b100}
\end{subfigure}
\begin{subfigure}[b]{0.33\linewidth}
\centering
\includegraphics[width=\textwidth]{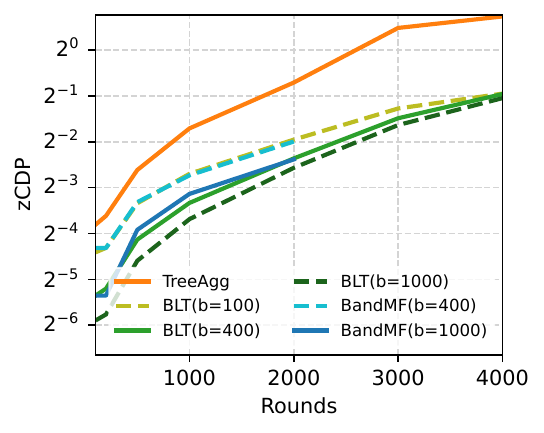}
\caption{\small Min-sep b=400}
\label{fig:varyn_b400} 
\end{subfigure}
\begin{subfigure}[b]{0.33\linewidth}
\centering
\includegraphics[width=\textwidth]{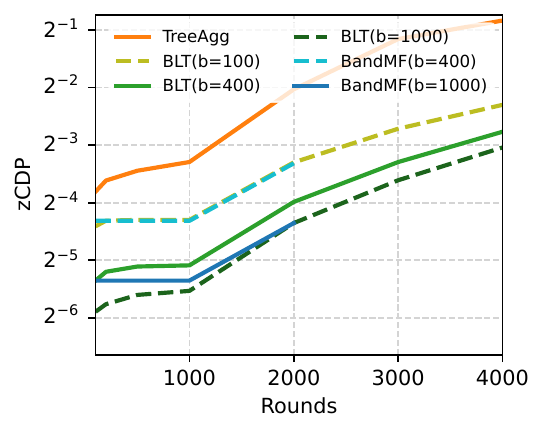}
\caption{\small Min-sep b=1000}
\label{fig:varyn_b1000} 
\end{subfigure}
\caption{\small Extrapolate by varying number of rounds $n$ for the \treeagg, \blt and \bandmf mechanisms used in production. Use the noise multiplier for the fixed report goal $6500$; fix min separation $b=100,400,1000$, respectively; worst-case max participation is varied assuming fixed population size, i.e., $k= n / b$. The utility of different mechanisms at a specific round (x-axis value) are assumed to be similar due to the signal-to-noise ratio hypothesis, and we can compare the corresponding zCDP guarantees (y-axis value).}\label{fig:varyn}
\end{figure}

Finally, in \cref{fig:varyn}, we extrapolate the DP guarantee results by varying the number of total rounds $n$ with the noise multiplier for the fixed report goal $6500$, fixed min separation $b=100,400,1000$, and corresponding max participation $k= n / b$. The \treeagg, \blt and \bandmf mechanisms used in production are compared. Instead of using \rmsloss or \maxloss to measure privacy-utility trade-offs in \cref{fig:rmse,fig:rmse_vary_n}, here we fix utility based on empirical utility of the production training and the signal-to-noise-ratio hypothesis, and compare the DP guarantees. As mentioned before, using \bandmf beyond the optimized matrices for $n=2000$ has not been studied before, and hence we only extrapolate \bandmf up to $n=2000$ rounds. \treeagg and \blt can run arbitrary number of rounds, and \blts achieve stronger DP guarantees than \treeagg. In practice, we can use one of the BLTs as a default mechanism across different settings, and perform on-the-fly optimization for given customized setting.  

\paragraph{Model launches} We have launched several language models for on-the-fly (OTF) rescoring words in Gboard smart completion and suggestion~\citep{gboard_dp_blogpost,xu2023federated}. The models trained with BLT-DP-FTRL algorithm in this paper achieved better privacy-utility trade-off than the \treeagg-DP-FTRL algorithm~\citep{kairouz21practical} widely used in FL practice~\citep{xu2023federated}, and comparable with MF-DP-FTRL~\citep{choquette2023amplified} that achieves state-of-the-art performance and some $\epsilon<1$ models reported in~\citep{gboard_dp_blogpost}. We report $\rho$-zCDP and $(\epsilon, \delta)$-DP of $\delta=10^{-10}$ following~\citet{xu2023federated,gboard_dp_blogpost}; the details of the DP configuration are provided in \cref{app:privacy-guarantees}. The Portuguese OTF LM in Brazil (pt-BR) is launched with DP $\epsilon=1.234$ (alternatively, zCDP $\rho=0.022$);
pt-PT OTF LM is launched with DP $\epsilon=8.19$ (alternatively, zCDP $\rho=0.749$) compared to DP $\epsilon=9.56$ (alternatively, zCDP $\rho=0.99$) in~\citep{xu2023federated}, and significant improvement in WMR of $-0.91\%$; 
es-ES OTF LM is launched with DP $\epsilon=4$ (alternatively, zCDP $\rho=0.202$) compared to DP $\epsilon=9.01$ (alternatively, zCDP $\rho=0.89$) in~\citep{xu2023federated}; the id-ID OTF LM using new vocabulary constructed from~\citep{gboard_fa_blogpost} is launched with DP $\epsilon=2.74$ (alternatively, zCDP $\rho=0.099$), achieving comparable utility and replacing the previous model without DP. In addition, the discussed default BLT mechanism optimized for $b=400$ and noise multiplier 7.379 is used to train and launch more models: the English LM is launched with DP $\epsilon=4.84$ (alternatively, zCDP $\rho=0.29$); the Spanish LM in Argentina is launched with DP $\epsilon=5.73$ (alternatively, zCDP $\rho=0.392$); the Spanish LM in Mexico is launched with DP $\epsilon=11.58$ (alternatively, zCDP $\rho=1.39$); the Spanish LM in other Latin America countries is launched with  DP $\epsilon=6.84$ (alternatively, zCDP $\rho=0.54$). The new Gboard OTF LMs are trained with FL and DP following the commitment in \citep{gboard_dp_blogpost,zhang2023private}. As of May 2025, all existing FL LMs have been replaced with DP FL models, fully realizing our plan in~\citep{zhang2023private}. 

%% file: fig_extrap.tex
\begin{figure}[thb]
\centering
\begin{subfigure}[b]{0.45\linewidth}
\centering
\includegraphics[width=\textwidth]{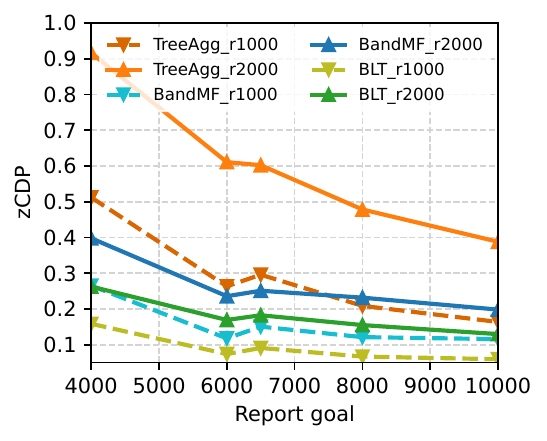}
\caption{\small Population 3M}
\label{fig:pop3m}
\end{subfigure}
\begin{subfigure}[b]{0.45\linewidth}
\centering
\includegraphics[width=\textwidth]{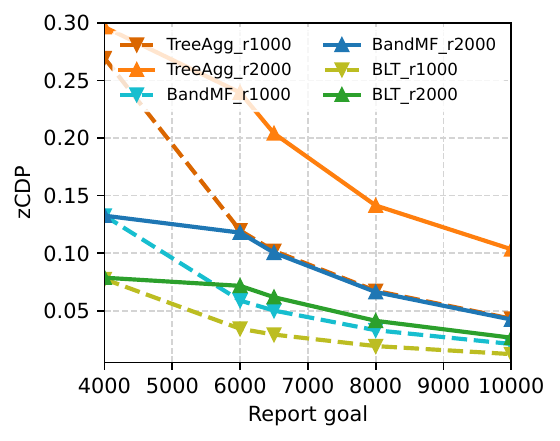}
\caption{\small Population 10M}
\label{fig:pop10m}
\end{subfigure}
\begin{subfigure}[b]{0.45\linewidth}
\centering
\includegraphics[width=\textwidth]{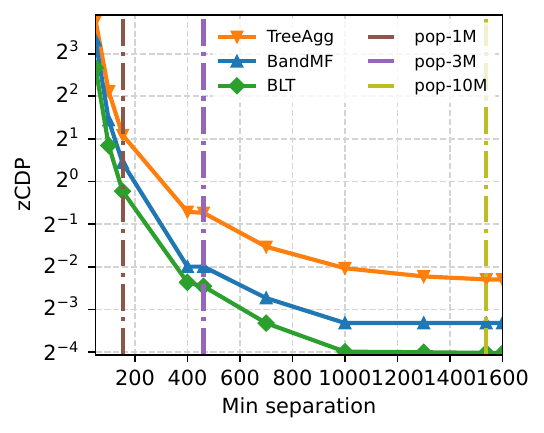}
\caption{\small Report goal 6500, worst-case max-par}
\label{fig:min_sep}
\end{subfigure}
\begin{subfigure}[b]{0.45\linewidth}
\centering
\includegraphics[width=\textwidth]{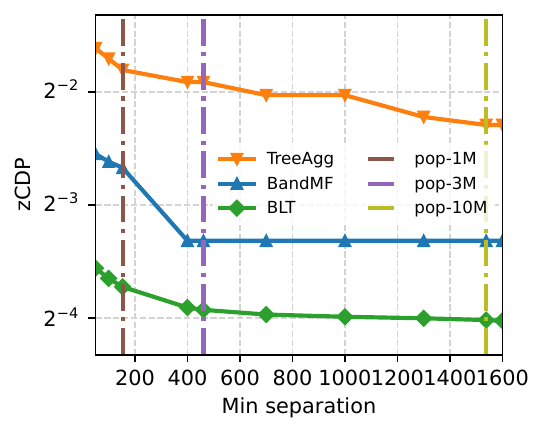}
\caption{\small Report goal 6500, optimistic max-par=2}
\label{fig:min_sep_max2}
\end{subfigure}
\caption{\small The effect of population size, number of rounds, report goals, and min-seps on DP-FTRL privacy guarantees. The results are extrapolate from the setting for es-ES and id-ID (\bandmf and \blt matrices are optimized for min-sep=400) based on the hypothesis that linearly scale noise multiplier and report goal, or only change min-sep will not affect the model utility. The For a fixed number of rounds to achieve utility target, increasing report goal and min-sep can achieve stronger guarantees measured by smaller zCDP. The optimal min-sep is capped by population size for a fixed report goal, and \blt provides better guarantees, and smoother transition across different min-seps. 
}\label{fig:extrapolate}
\end{figure}

%% file: sec_concl.tex
\section{Concluding Remarks}
\vspace{-0.2cm}
This work addresses the critical challenge of achieving strong DP in FL for on-device LMs. We have successfully extended the BLT mechanism to multi-participation scenarios and integrated it into the DP-FTRL framework. Our BLT-DP-FTRL algorithm demonstrates superior privacy-utility trade-offs compared to the widely-used \treeagg mechanism while maintaining its ease of use. Furthermore, it rivals the state-of-the-art \bandmf mechanism in performance, yet without the associated complexities and high memory costs. Through extensive empirical evaluations on both a benchmark dataset and real-world production tasks, we have showcased the practicality and effectiveness of BLT-DP-FTRL, paving the way for its broader adoption. 

The empirical results in this paper primarily focus on the cross-device FL setting where privacy amplification by sampling is challenging in practice. The discussions (e.g. \cref{tab:mechanism_summary}) can also be applied to centralized setting for user-level DP or example-level DP. In centralized setting, \bandmf~\citep{choquette2023amplified} with amplification can achieve better privacy-utility trade-off measured by \rmsloss among the mentioned mechanisms, when number of rounds $n$ and model dimension $m$ is not too large for optimizing and applying the mechanism. When $n$ and $m$ are large, \blt and \bandtoep~\citep{mckenna2024scaling} (similarly, \bandfhu~\citep{kalinin2024banded}) can both be applied, where \blt has less optimization cost for very large $n$ (shown in \cref{fig:faster}), while \bandtoep can apply the known amplification by sampling mechanism. In conclusion, BLT-DP-FTRL algorithm is recommended for cross-device federated learning~\citep{xu2023federated,kairouz2019advances}, and is also useful for DP in datacenter~\citep{ponomareva2023dpfy,xu2022learning,charles2024fine} and federated learning in trusted execution environments (TEEs)~\citep{daly2024federated,eichner2024confidential}.

%% file: sec_app.tex
\ifthenelse{\boolean{acl}}{
\section{Additional Background on Federated Learning (FL) with Differential Privacy (DP)} \label{app:background}
\input{subsec_dp_def}
\subsection{DP-FTRL for DP FL algorithm}
\input{alg_dpfl}
\subsection{\treeagg and \bandmf in DP-FTRL} \label{app:ta_mf}
\input{subsec_ta_mf}
}{}

\ifthenelse{\boolean{acl}}{
\section{Stream Multiplication by BLT matrices $C$ and $C^{-1}$}
\input{alg_stream_multi}
}{}

\ifthenelse{\boolean{acl}}{
\section{More discussion on BLT-DP-FTRL} \label{app:blt-dp-ftrl}
\input{tab_mech}
\subsection{Background on Multi-participation Sensitivity} \label{app:mp_sens}
\input{subsec_blt_sens}
\subsection{A    Sensitivity    Lower    Bound} \label{app:sens_lower_bound}
\input{subsec_sens_lower_bound}
}{}

\ifthenelse{\boolean{acl}}{
\section{Optimizing for BLT Matrices}

\input{alg_opt_blt}
\input{subsec_opt_eff}
}{}

\ifthenelse{\boolean{acl}}{
\section{More \rmsloss and \maxloss Experiments} \label{app:mse}
\input{subsec_mse}
}{}

\section{BLT Parameters for Production Training} \label{app:blt_params}
We provide the \ourblt parameters we generated and used in training production LMs with DP FL in \cref{sec:gboard}. The \ourblt matrices are optimized for three min-sep settings $b=(100, 400, 1000)$ and each BLT is parameterized by 8 values for buffer size $d=4$, i.e., buffer decay $\theta \in \R^d$ and output scale $\omega \in \R^d$. 
\begin{itemize}
    \item min-sep $b=100$, total rounds $n=2000$, max participation $k=10  \, \Rightarrow$, \\
    $\theta=(        0.989739971007307,
        0.7352001759538236,
        0.16776199983448145,
        0.1677619998016191)$, 
        $\omega = (        0.20502892852480875,
        0.23357939425278557,
        0.03479503245420878,
        0.03479509876050538)$.
    \item min-sep $b=400$, total rounds $n=4000$, max participation $k=5  \, \Rightarrow$, \\
    $\theta=(0.9999999999921251,
        0.9944453083640997,
        0.8985923474607591,
        0.4912001418098778)$, 
        $\omega = (0.0070314825502323835,
        0.10613806907600574,
        0.1898159060327625,
        0.1966594748073734)$.
    \item min-sep $b=1000$, total rounds $n=4000$, max participation $k=2  \, \Rightarrow$, \\
    $\theta=(        0.9999999999983397,
        0.9973412136664378,
        0.9584629472313878,
        0.6581796870749317)$, 
        $\omega = (        0.008657392263671862,
        0.05890891298180163,
        0.14548176930698697,
        0.2770117005326523)$.
\end{itemize}

\begin{figure}[thb]
\centering
\begin{subfigure}[b]{0.45\linewidth}
\centering
\includegraphics[width=\textwidth]{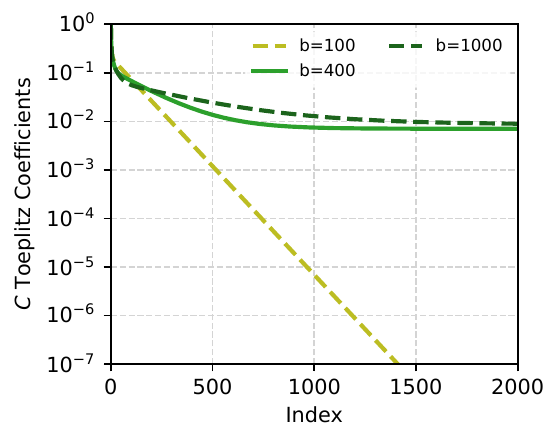}
\caption{\small }
\label{fig:blt_c}
\end{subfigure}
\begin{subfigure}[b]{0.44\linewidth}
\centering
\includegraphics[width=\textwidth]{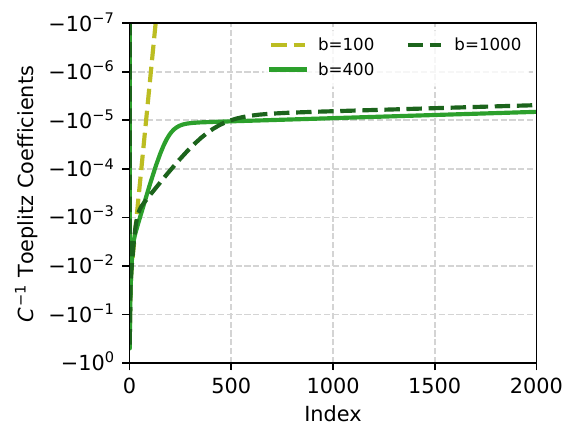}
\caption{\small }
\label{fig:blt_c_inv} 
\end{subfigure}
\caption{\small Toeplitz coefficients $\{c_0,\ldots,c_n\}$ for $\bfC=\LtToep(c)$ and $\{\hat c_0,\ldots,\hat c_n\}$ for $\bfC^{-1}=\LtToep(\hat c)$; For $\blt(\theta, \omega)$, $c$ can be computed by \cref{eq:closedcoefs}, and there are many ways to derive corresponding $\hat c$ (e.g., set $\bfZ=\bfI$ in \cref{alg:multCinv}).}\label{fig:blt_coeff}
\end{figure}

\cref{fig:blt_coeff} visualizes the corresponding Toeplitz coefficients for $\bfC$ to compute sensitivity and $\bfC^{-1}$ for generating correlated noise. The coefficients of $\blt(\theta, \omega)$ for $b=100$ decaying faster than $b=400$ and $b=1000$. 

\section{Additional Simulation and Production Results} \label{app:plots}
\ifthenelse{\boolean{acl}}{
\subsection{Production Setting for Mobile Keyboard LMs} \label{app:prod_config}
\input{subsec_gboard_config}
\subsection{Extrapolation Results for Production Setting} \label{app:extrap}
\input{subsec_gboard_extrap}
\subsection{Additional Plots}
}{}

\begin{figure*}[tbh]
    \centering
    \includegraphics[width=0.8\linewidth]{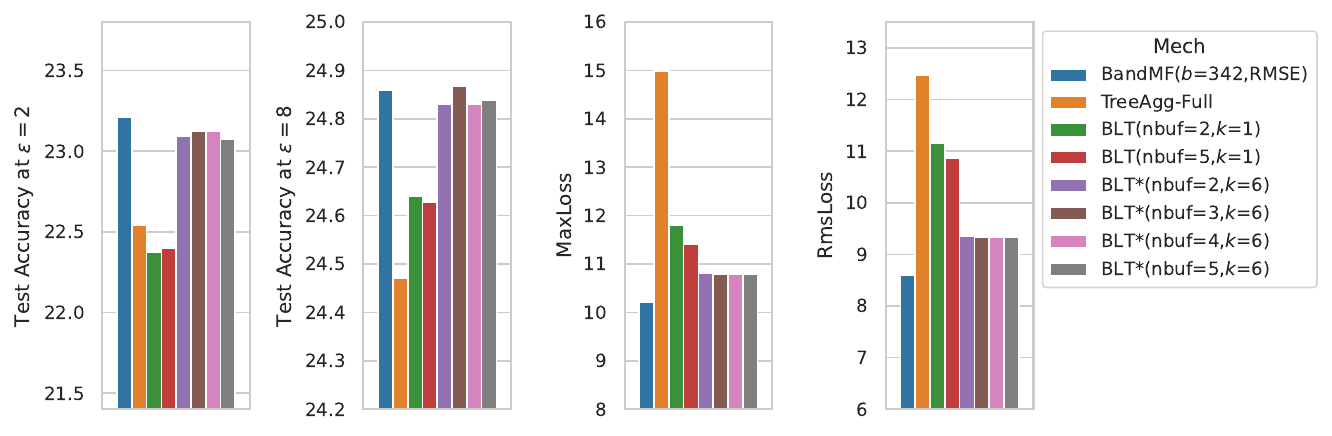}
    \caption{\small Visualizing the results in \cref{tab:sonwp}.}
    \label{fig:sonwp}
\end{figure*}

\begin{figure*}[tbh]
    \centering
    \includegraphics[width=\linewidth]{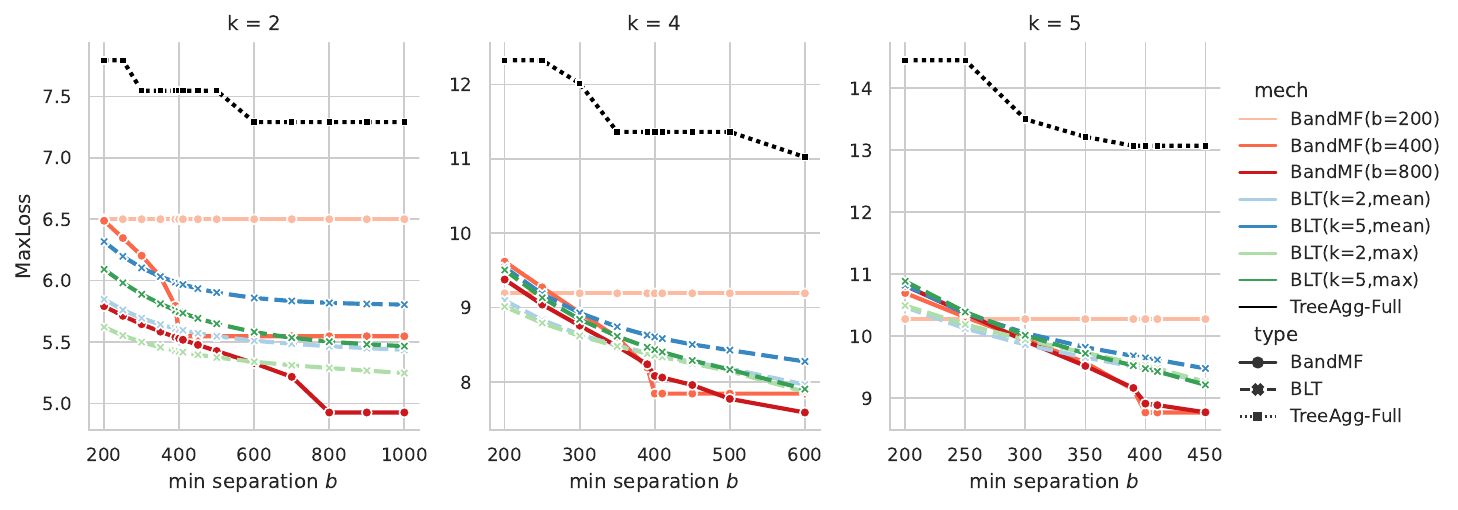}
    \caption{\small The same mechanisms from \cref{fig:rmse}, but compared on \maxloss instead of \rmsloss.}
    \label{fig:maxloss}
\end{figure*}

\input{fig_acc}
\input{fig_acc2}
\input{fig_priv}
\clearpage
\section{Reporting privacy guarantees}\label{app:privacy-guarantees}
This section clarifies the nuances of the DP guarantees following the guidelines outlined in \citep[Sec. 5.3]{ponomareva2023dpfy}. They are similar (if not identical)  to \citep[Appendix A]{xu2023federated} and \citep[Appendix C.3]{wu2024prompt}, except for the exact DP guarantees. We include the entire checklist for completeness.

\begin{enumerate}
    \item \textbf{DP setting}. This is a central DP guarantee where the service provider is trusted to correctly implement the mechanism.
    \item \textbf{Instantiating the DP Definition}
     \begin{enumerate}
        \item \textit{Data accesses covered}: The DP guarantee applies to all well-behaved clients\footnote{Clients that faithfully follow the algorithm including participation limits. Due to the design of the algorithm, a mis-behaved client does not adversely affect the DP guarantee of any well-behaved clients.} in a single training run. We do not account for hyperparameter tuning, or the selection of the final model checkpoint using evaluation metrics or A/B testing in our guarantees. Public data such as C4~\citep{xue2020mt5} or LLM-based synthetic data~\citep{wu2024prompt}, are used for pre-training. 
        \item \textit{Final mechanism output}: Only the final model checkpoint is released for production deployment, however the mechanism’s output is technically the full sequence of privatized gradients, and so the guarantee also applies at this level, and hence all intermediate models are protected (including those sent to devices participating in federated learning). 
        \item \textit{Unit of privacy}. Device-level DP is considered, i.e., the notion of adjacency is with respect to arbitrary training datasets on each client device, and the device might have an arbitrarily large local dataset containing arbitrary training examples. For user's with a single device, this corresponds directly to user-level DP; for devices shared with multiple users, this provides a stronger notion of DP than user-level; for a user with multiple devices that happen to both participate in training the model, the notion is weaker, but group privacy can be used to obtain a user-level guarantee.
        \item \textit{Adjacency definition for ``neigbouring'' datasets}: We use the zero-out definition~\citep{kairouz21practical}. This is a a special form of the add-or-remove definition, where neighboring data sets differ by addition/removal of a single client. In the absence of a client at any training step, we assume that the client's model update gets replaced with the all zeros vector. This assumption enforces a subtle modification to the traditional definition of the add/remove notion of DP which allows neighboring data sets to have the same number of records.
    \end{enumerate}
    \item \textbf{Privacy accounting details}
    \begin{enumerate}
        \item \textit{Type of accounting used}: Both $\rho-$zCDP~\citep{bun2016concentrated} accounting, and PLD accounting~\citep{pldlib} for $(\epsilon, \delta)-$DP are used.
        \item \textit{Accounting assumptions }: Each client only participates limited times during the training, and there are at least a min-separation number of rounds between two consecutive participation of a client. Client participation is enforced by a timer on clients in the cross-device FL system.    
        \item \textit{The formal DP statement}: See \cref{sec:gboard} \textbf{model launches} for reported DP guarantees.
        \item \textit{Transparency and verifiability}: We use the open-sourced core implementation code in TensorFlow Federated and Tensorflow Privacy. Key portions of the cross-device FL system are also open sourced.
    \end{enumerate}
\end{enumerate}

%% file: fig_acc.tex
\begin{figure*}[thb]
\centering
\begin{subfigure}[b]{0.45\linewidth}
\centering
\includegraphics[width=\textwidth]{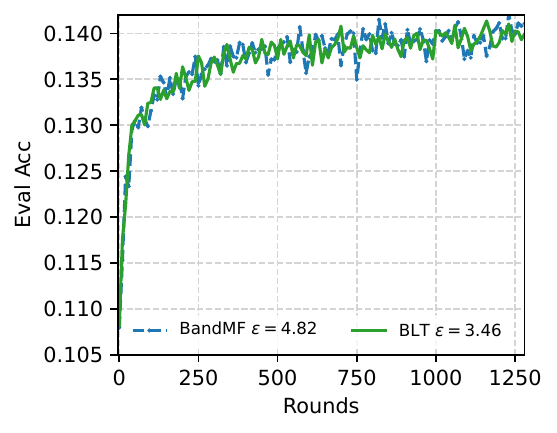}
\caption{\small Spanish LM in Spain (es-ES)}
\label{fig:es_acc}
\end{subfigure}
\begin{subfigure}[b]{0.45\linewidth}
\centering
\includegraphics[width=\textwidth]{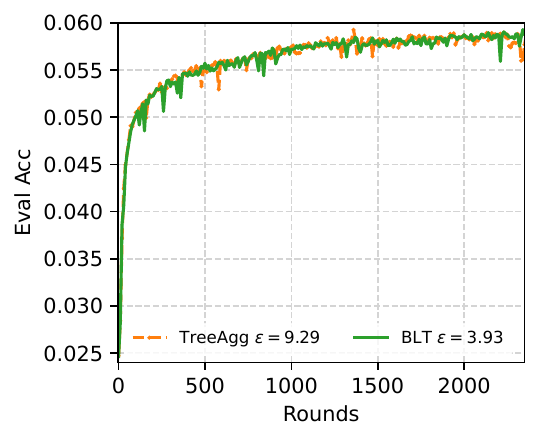}
\caption{\small Indonesian LM in Indonesia (id-ID)}
\label{fig:id_acc}
\end{subfigure}
\begin{subfigure}[b]{0.45\linewidth}
\centering
\includegraphics[width=\textwidth]{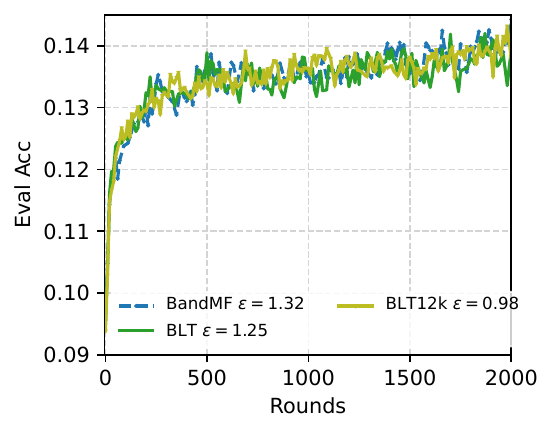}
\caption{\small Portuguese LM in Brazil (pt-BR)}
\label{fig:br_acc}
\end{subfigure}
\begin{subfigure}[b]{0.45\linewidth}
\centering
\includegraphics[width=\textwidth]{images/ptpt_acc.pdf}
\caption{\small Portuguese LM in Portugal (pt-PT)}
\label{fig:pt_acc}
\end{subfigure}
\caption{\small The NWP evaluation accuracy curves for training LMs with DP-FTRL in FL. \blt achieves comparable NWP accuracy and slightly better privacy guarantees (at the last round) compared to \bandmf for es-ES and pt-BR; much better DP guarantees, and/or better utility compared to \treeagg for id-ID and pt-BR.
}\label{fig:acc_curve}
\end{figure*}

%% file: fig_acc2.tex
\begin{figure*}[thb]
\centering
\begin{subfigure}[b]{0.45\linewidth}
\centering
\includegraphics[width=\textwidth]{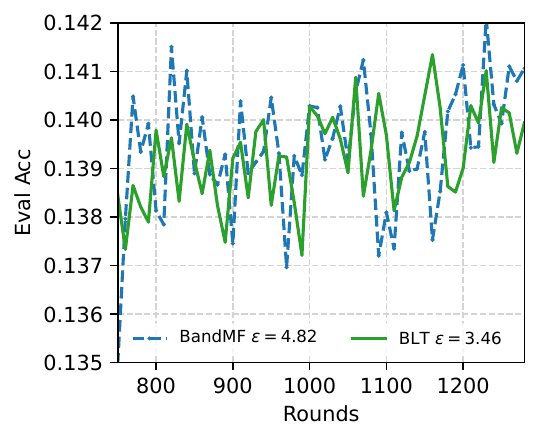}
\caption{\small  Spanish LM in Spain (es-ES)}
\label{fig:es_acc2}
\end{subfigure}
\begin{subfigure}[b]{0.45\linewidth}
\centering
\includegraphics[width=\textwidth]{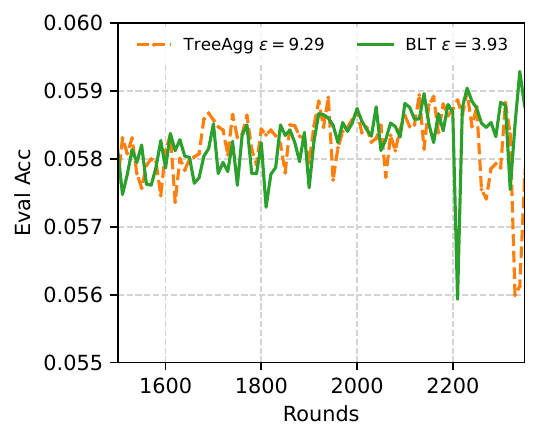}
\caption{\small  Indonesian LM in Indonesia (id-ID)}
\label{fig:id_acc2}
\end{subfigure}
\begin{subfigure}[b]{0.45\linewidth}
\centering
\includegraphics[width=\textwidth]{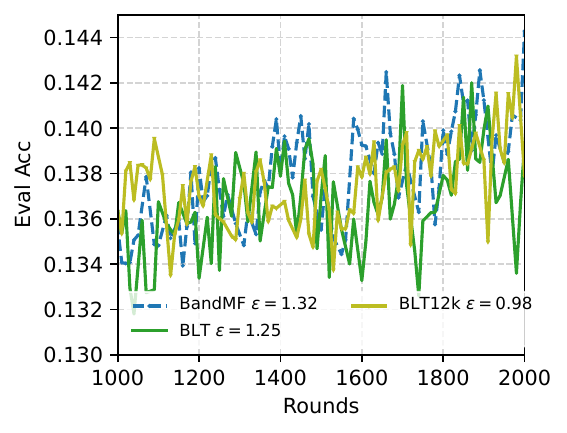}
\caption{\small  Portuguese LM in Brazil (pt-BR)}
\label{fig:br_acc2}
\end{subfigure}
\begin{subfigure}[b]{0.45\linewidth}
\centering
\includegraphics[width=\textwidth]{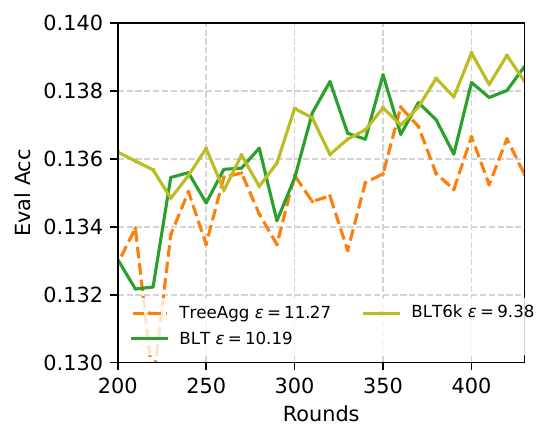}
\caption{\small  Portuguese LM in Portugal (pt-PT)}
\label{fig:pt_acc2}
\end{subfigure}
\caption{\small  The NWP evaluation accuracy curves for training LMs with DP-FTRL in FL. Zoom in the latter stage of training for the curves in \cref{fig:acc_curve}. The NWP accuracy increases fast in the first ~200 rounds in DP FL training, and the accuracy changes within the range of $0.01$ when zooming in the later stage. The oscillation is because of the stochasticity in forming subsets of devices in both training and evaluation per round. The average NWP accuracy from nearby rounds is reported in \cref{tab:gboard} to reduce the variance. }\label{fig:acc_curve2}
\end{figure*}

%% file: fig_priv.tex
\begin{figure*}[thb]
\centering
\begin{subfigure}[b]{0.45\linewidth}
\centering
\includegraphics[width=\textwidth]{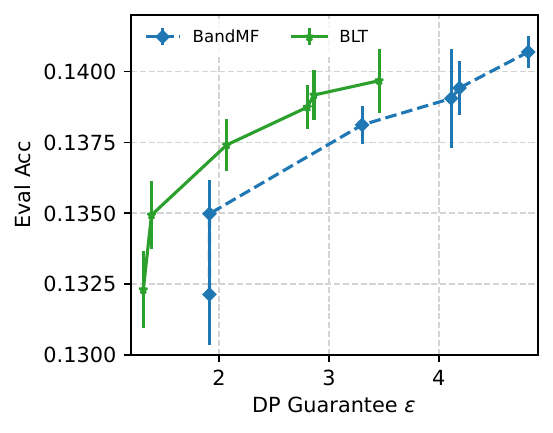}
\caption{\small Spanish LM in Spain (es-ES)}
\label{fig:es_priv}
\end{subfigure}
\begin{subfigure}[b]{0.45\linewidth}
\centering
\includegraphics[width=\textwidth]{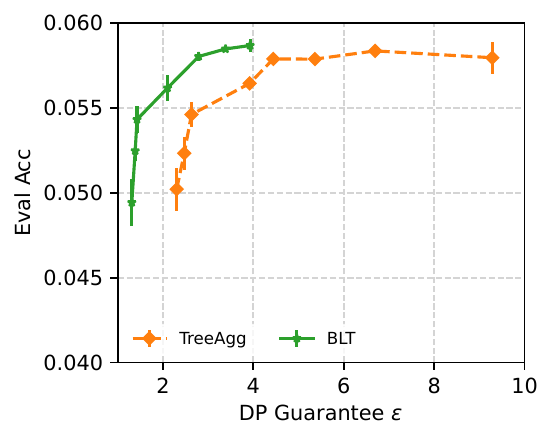}
\caption{\small Indonesian LM in Indonesia (id-ID)}
\label{fig:id_priv}
\end{subfigure}
\begin{subfigure}[b]{0.45\linewidth}
\centering
\includegraphics[width=\textwidth]{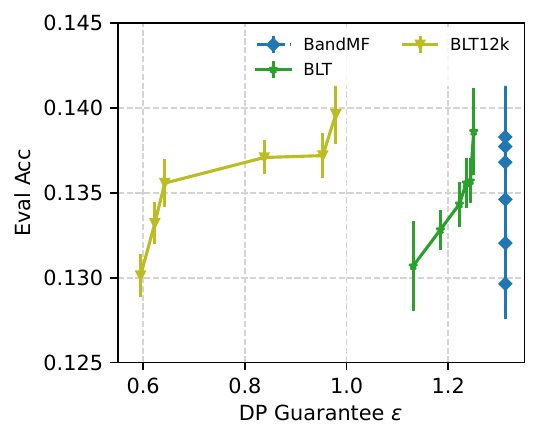}
\caption{\small Portuguese LM in Brazil (pt-BR)}
\label{fig:br_priv}
\end{subfigure}
\begin{subfigure}[b]{0.45\linewidth}
\centering
\includegraphics[width=\textwidth]{images/ptpt_priv.pdf}
\caption{\small Portuguese LM in Portugal (pt-PT)}
\label{fig:pt_priv}
\end{subfigure}
\caption{\small The privacy-utility trade-off curves derived from \cref{fig:acc_curve}. For each selected round $r$, we compute the mean and standard deviation (shown as vertical bars) for accuracy from the rounds in the range of $r\pm50$ ($r\pm10$ for pt-PT), and also accounting the DP guarantees. \blts show better privacy-utility trade-off as their curves are closer to the top left (small DP guarantees and large NWP accuracy).
}\label{fig:priv_curve}
\end{figure*}